\documentclass[preprint,12pt]{elsarticle}

\usepackage{amssymb}

\usepackage{lineno}

\usepackage[latin1]{inputenc}
\usepackage{hyperref}

\usepackage{amsmath, amssymb}
\usepackage{subcaption}

\usepackage{enumitem}

\usepackage{longtable}

\usepackage[table]{xcolor} 
\usepackage{colortbl}
\definecolor{light-gray}{HTML}{E5E4E2}

\usepackage{soul}

\newcommand{\R}{\mathbb{R}} 
\newcommand{\F}{\mathcal{F}} 
\newcommand{\Like}{\mathcal{L}} 
\newcommand{\Humans}{\Phi} 
\newcommand{\human}{\phi} 
\newcommand{\E}{\mathcal{E}} 
\newcommand{\Data}{\mathcal{D}} 
\newcommand{\N}{\mathcal{N}} 

\newcommand{\papertitle}{A Bayesian Evaluation Framework for Subjectively~Annotated~Visual~Recognition~Tasks}

\journal{Pattern Recognition}

\begin{document}

\begin{frontmatter}

\title{\papertitle}

\author{Derek S. Prijatelj\fnref{add1}}
\ead{dprijate@nd.edu}
\ead[https://prijatelj.github.io/]{prijatelj.github.io}

\author{Mel McCurrie\fnref{add2}}
\ead{mel@perceptiveautomata.com}

\author{Samuel E. Anthony\fnref{add2}}
\ead{santhony@perceptiveautomata.com}

\author{Walter J. Scheirer\fnref{add1}\corref{cor1}}
\cortext[cor1]{The corresponding author}
\ead{walter.scheirer@nd.edu}
\ead[https://www.wjscheirer.com/]{wjscheirer.com}

\address[add1]{Department of Computer Science and Engineering, University of Notre Dame, Notre Dame, IN 46556, USA}
\address[add2]{Perceptive Automata, One Boston Place, Suite 3520, 201 Washington Street, Boston, MA 02108, USA}

\begin{abstract}
    An interesting development in automatic visual recognition has been the emergence of tasks where it is not possible to assign objective labels to images, yet still feasible to collect annotations that reflect human judgements about them.
Machine learning-based predictors for these tasks rely on supervised training that models the behavior of the annotators, \textit{i.e.}, what would the average person's judgement be for an image? 
A key open question for this type of work, especially for applications where inconsistency with human behavior can lead to ethical lapses, is how to evaluate the epistemic uncertainty  of trained predictors, \textit{i.e.}, the uncertainty that comes from the predictor's model.
We propose a Bayesian framework for evaluating black box predictors in this regime, agnostic to the predictor's internal structure.
The framework specifies how to estimate the epistemic uncertainty that comes from the predictor with respect to human labels by approximating a conditional distribution and producing a credible interval for the predictions and their measures of performance.
The framework is successfully applied to four image classification tasks that use subjective human judgements: facial beauty assessment, social attribute assignment, apparent age estimation, and ambiguous scene labeling.

\end{abstract}


\end{frontmatter}

Graphical Abstract
\begin{center}
    \includegraphics[width=.7682\textwidth]{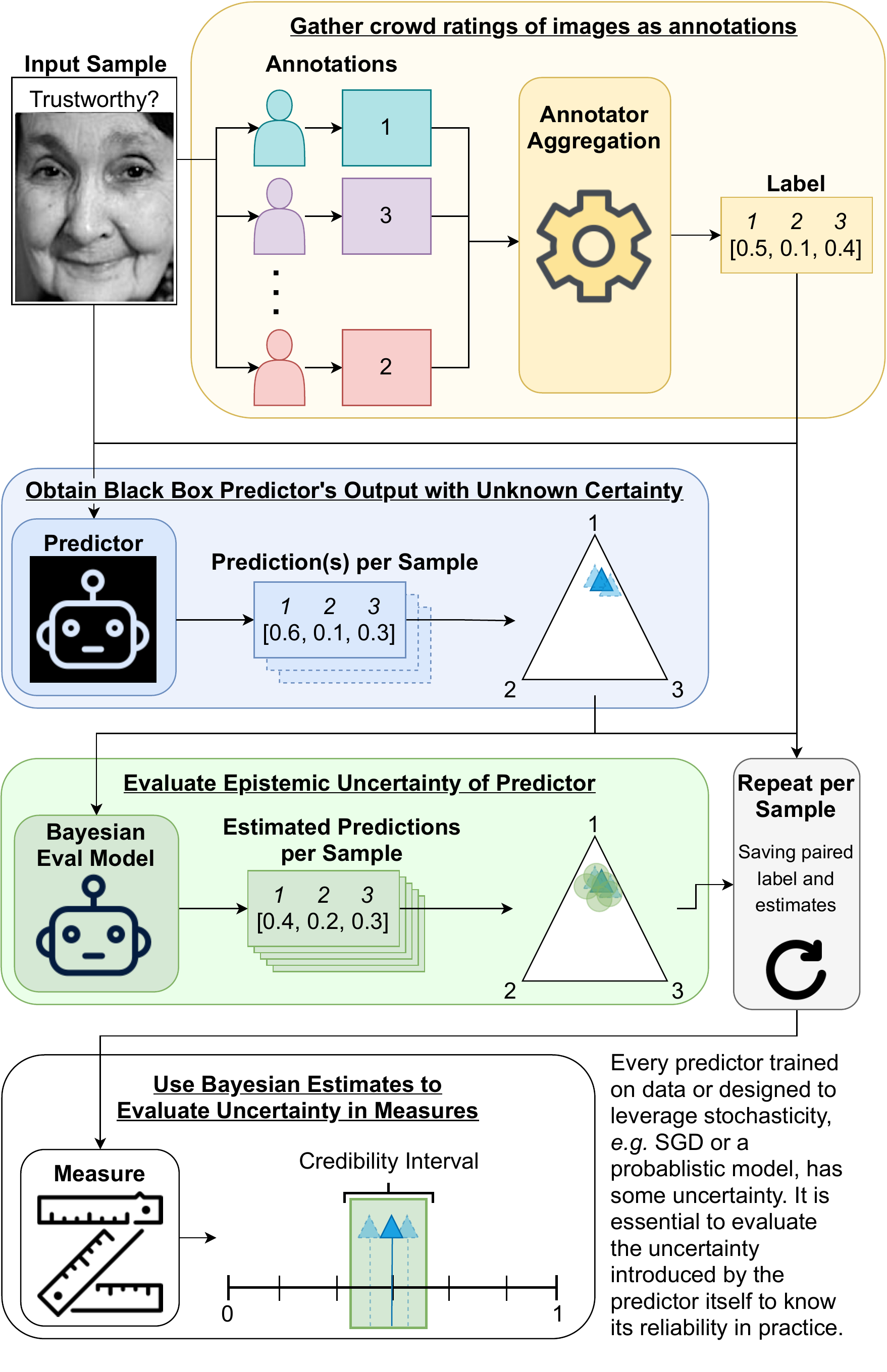}
\end{center}

\section{Introduction}
An intriguing application of machine learning is visual recognition when there is a lack of available objective truth to serve as the target labels, leaving noisy crowd judgements as an alternative.
These tasks are subjectively annotated visual recognition tasks such as crowd judgements of facial beauty
~\cite{liang_scut-fbp5500:_2018}, personality attributes from faces~\cite{mccurrie2017predicting, ponce2016chalearn}, and ambiguous scene labeling~\cite{rodrigues_learning_2017}.
The predictors for each of these tasks are trained to predict the expected responses of a crowd of annotators.
A problem arises in the assessment of how well a predictor matches the expected annotator behavior and its certainty, which is key to verifying the predictor's behavioral fidelity, and is important to both machine learning researchers
~\cite{escalera2017chalearn} and psychologists alike~\cite{todorov2013validation}.
For applications where ethical lapses can lead to unfortunate consequences, such predictors need special care to understand the characteristics of their output and to quantify their uncertainty.

To avoid such problems, a comprehensive evaluation framework that compares predictors to human data in a probabilistic manner is necessary.
Previous predictor evaluation in this setting used  summary statistics for the traditional supervised task~\cite{ponce2016chalearn, rodrigues_learning_2017}.
This regime is known to lead to unintended consequences in practice
~\cite{abdar_review_2021}.
Such statistics do not account for uncertainty in the predictor nor noisy labels obtained from a crowd of annotators.

In this work, we attempt to answer the following question, which directly addresses predictor uncertainty: ``Given an accepted label value $y$, what values and their probabilities does the predictor confuse with that value $y$?''
The proposed Bayesian Evaluation framework, as seen in the graphical abstract and Fig.~\ref{fig:framework}, includes a Bayesian Evaluator that takes as input a given annotator aggregation, and a predictor trained on features derived from training images using the annotator aggregation data as the target labels.
Annotator aggregation is a data design decision related to how to summarize and represent the human behavior to be modeled by the predictor
~\cite{zheng_truth_2017}
, \textit{e.g.}, human frequency in annotations as the label.
After using that input data to obtain a generative model for the evaluator, the Bayesian Evaluator may be sampled.
The samples serve as draws from the conditional distribution, which indicate the predictor's confusion for a given annotator aggregated label.
The conditional distribution of the predictor's output can then be used to obtain the distribution of measures by applying the desired measure~\cite{tao2011introduction} to the human-derived label and to the draws of the conditional distribution given that label.
This measure distribution can subsequently be used to assess uncertainty by defining a credible interval about the measure,
such as the highest posterior density interval (HPDI)
~\cite{ohagan_kendalls_2004}
[SM~2.2.1].

If the predictor is non-stochastic, then multiple predictions representative of the predictor may need to be obtained to provide a stronger source of information of the predictor's uncertainty, \textit{e.g.}, multiple instances of the predictor on the same fold or across different folds, or through drop-out at test time~\cite{gal_uncertainty_2016} in the case of artificial neural networks (ANN).
If representative samples of each prediction per input is not available, then the generative model's design and any used prior information will will have a greater impact in determining the predictor's epistemic uncertainty.

\textbf{Contributions.}
\textbf{a}) We define a Bayesian evaluation framework for obtaining the conditional probability distribution of a predictor's output given the annotator aggregated target labels, $P_{\hat{Y}}(\hat{Y}|Y)$, which, as far as the authors know, has not been done to date in epistemic uncertainty estimators nor in calibration of predictors
~\cite{abdar_review_2021}.
\textbf{b}) We describe and demonstrate two implementations of that framework, one via a statistical model and another with a Bayesian Neural Network (BNN).
\textbf{c}) The proposed framework is used to evaluate the epistemic uncertainty of black box predictors on four image classification tasks that use subjective human judgements, including facial beauty assessment using the SCUT-FBP5500 dataset~\cite{liang_scut-fbp5500:_2018}, social attribute assignment using data from TestMyBrain.org~\cite{germine2012web}, apparent age estimation using data from the ChaLearn series of challenges~\cite{escalera2017chalearn}, and ambiguous scene labeling using the LabelMe dataset~\cite{russell2008labelme}.
Our code for the proposed framework will be made available online after publication.
The Supplementary Material (SM) is provided at
\url{https://arxiv.org/abs/2007.06711}
for more background information and implementation details.
The notation used is described in detail in SM~2.1.

\section{Background and Related Work} 
\label{sec:background}

\textbf{Annotator Aggregation.}
Given multiple annotations per sample, then various annotator aggregation methods apply that address annotator reliability and how to extract the task-relevant signal from their annotations~\cite{zheng_truth_2017, hung2013evaluation}.
\cite{welinder2010multidimensional} suggested a model that combines the image formation and annotation process of crowds to predict ground truth labels.
\cite{ipeirotis_quality_2010} modeled bias and error in groups annotating samples on Amazon's Mechanical Turk service.
\cite{rodrigues_deep_2018} proposed an EM algorithm for jointly learning the parameters of a neural network and the reliabilities of the annotators.
Subsequently, a general-purpose crowd layer for the network is proposed, which allows for end-to-end training from the noisy labels of multiple annotators.
All of these approaches are designed with an objective notion of ground truth in mind, and with a general assumption that it is obscured by noise. 

Soft labels can improve predictor performance in supervised learning by including the annotators' uncertainty~\cite{peterson_human_2019}.
Soft labels are the expected probabilities per class of the discrete distribution (\textit{i.e.}, a mean without variance) and while derived from data containing the aleatoric uncertainty (differences in human labeling behavior), they do not capture the aleatoric uncertainty of the probability vectors themselves, only the nominal values of the discrete distribution.
We trained using human frequency as soft labels, however we take this a step further by modeling the conditional distribution $P_{\hat{Y}}(\hat{Y} | Y)$, which is measure-agnostic and able to capture the predictor's epistemic uncertainty.
This means the framework introduced in this paper focuses only on the epistemic uncertainty of the predictor, while using the frequency of human labels.
Moreover, we perform this on subjectively annotated tasks.

When considering uncertainty, prior work considers aleatoric uncertainty and epistemic uncertainty as uncertainty inherent in the data and predictors respectively.
The uncertainty estimated and attempted to be removed in annotator aggregation is aleatoric uncertainty~\cite{abdar_review_2021}. 
This source of uncertainty may or may not be carried through to the final trained predictor.
Regardless, the way a predictor learns during training will typically induce some amount of epistemic uncertainty, which is what this paper focuses on.

\textbf{Epistemic Uncertainty and Calibration.}
Work to address epistemic uncertainty has appeared recently
~\cite{abdar_review_2021}, including works on Gaussian Discriminant Classifiers~\cite{carranza_alarcon_imprecise_2021}, the task of visual-dialog~\cite{patro_probabilistic_2021}, and
multi-target regression
~\cite{messoudi_copula-based_2021}.
The evaluation of epistemic uncertainty is often calculated as the calibration function in the calibration of classifiers and regressors
~\cite{widmann2020calibration}.
The calibration function is the difference between the accepted label and prediction to assess a predictor's calibration, often desired to perform the remapping of miscalibrated predictions
~\cite{abdar_review_2021}.
However the conditional probability distribution of interest in calibration is $P_Y(Y=\hat{Y}|X)$, rather than $P_{\hat{Y}}(\hat{Y}|Y)$, which our framework focuses on.
See how they relate in SM~3.4.
Our framework relates to the Bayesian evaluation of black-box classifiers
~\cite{ji_label_efficient_2020}
, however our work models $P_{\hat{Y}}(\hat{Y}|Y)$ and is agnostic of task type.

\begin{figure*}[t]
    \centering
    \subcaptionbox{
    }[.31\textwidth]{
        \includegraphics[width=.32\textwidth]{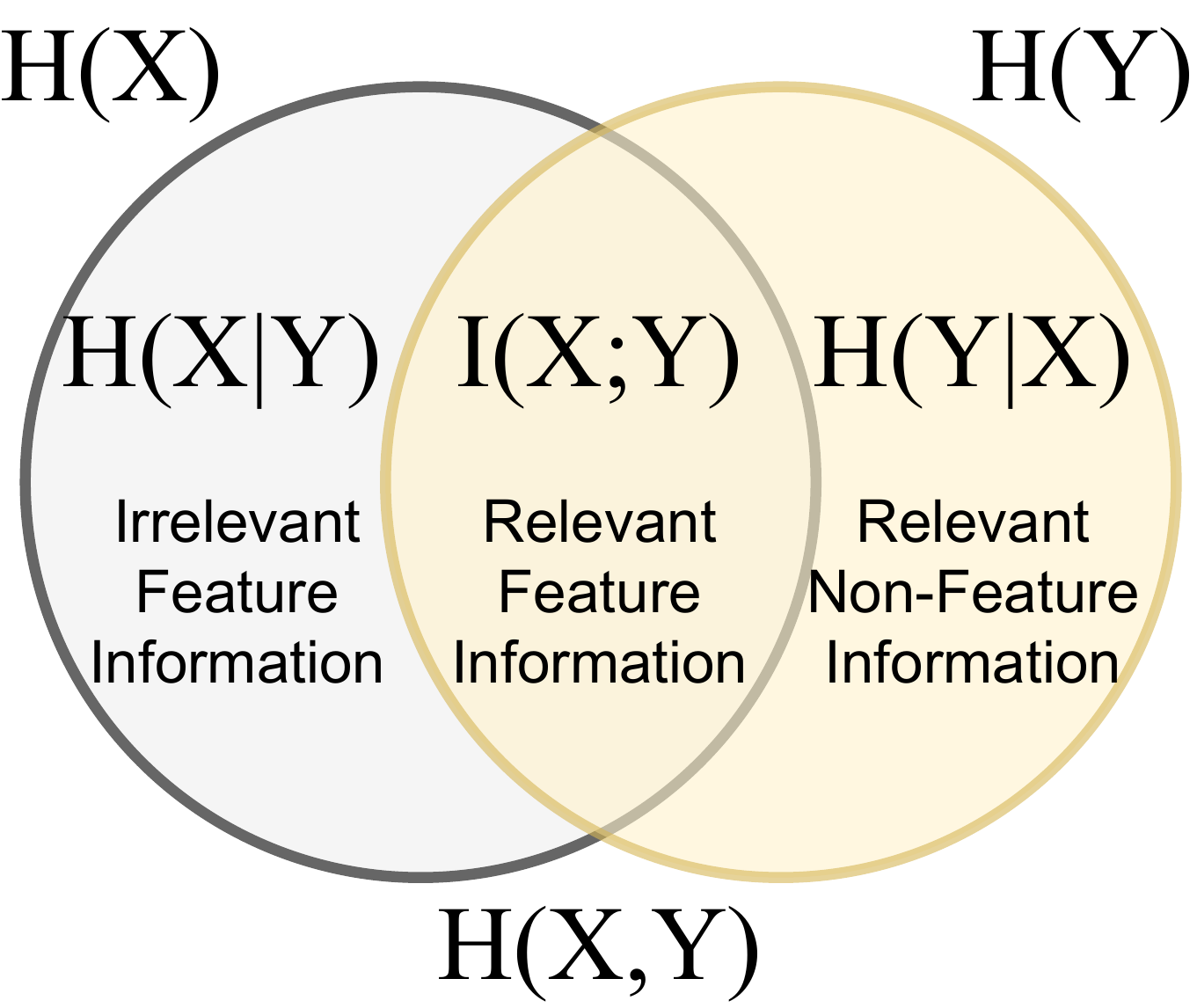}
        \label{fig:uncertainty_venn_xy}
    }
    \hfill
    \subcaptionbox{
    }[.31\textwidth]{
        \includegraphics[width=.32\textwidth]{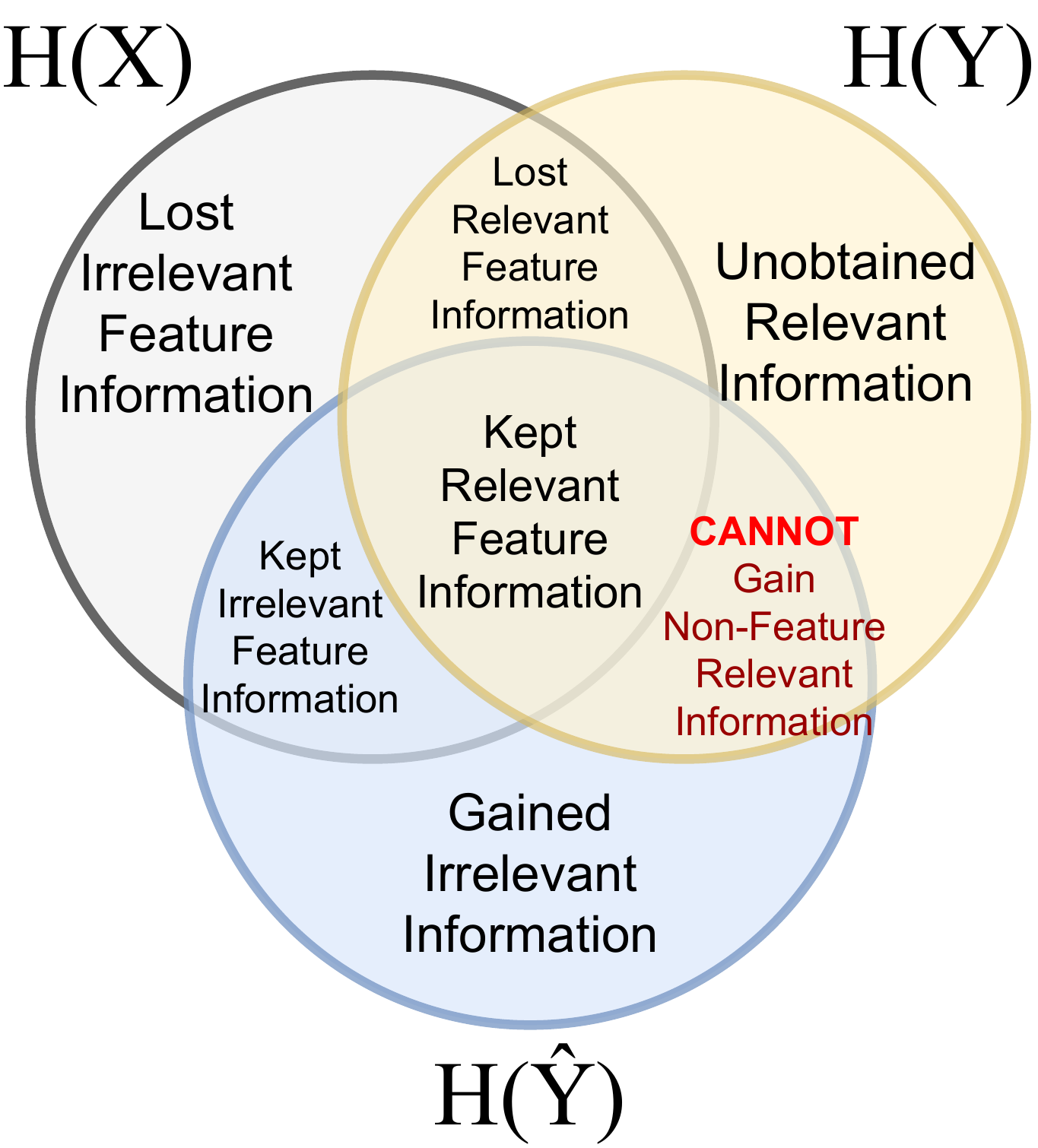}
        \label{fig:uncertainty_venn_xyyhat}
    }
    \hfill
    \subcaptionbox{
    }[.31\textwidth]{
        \includegraphics[width=.32\textwidth]{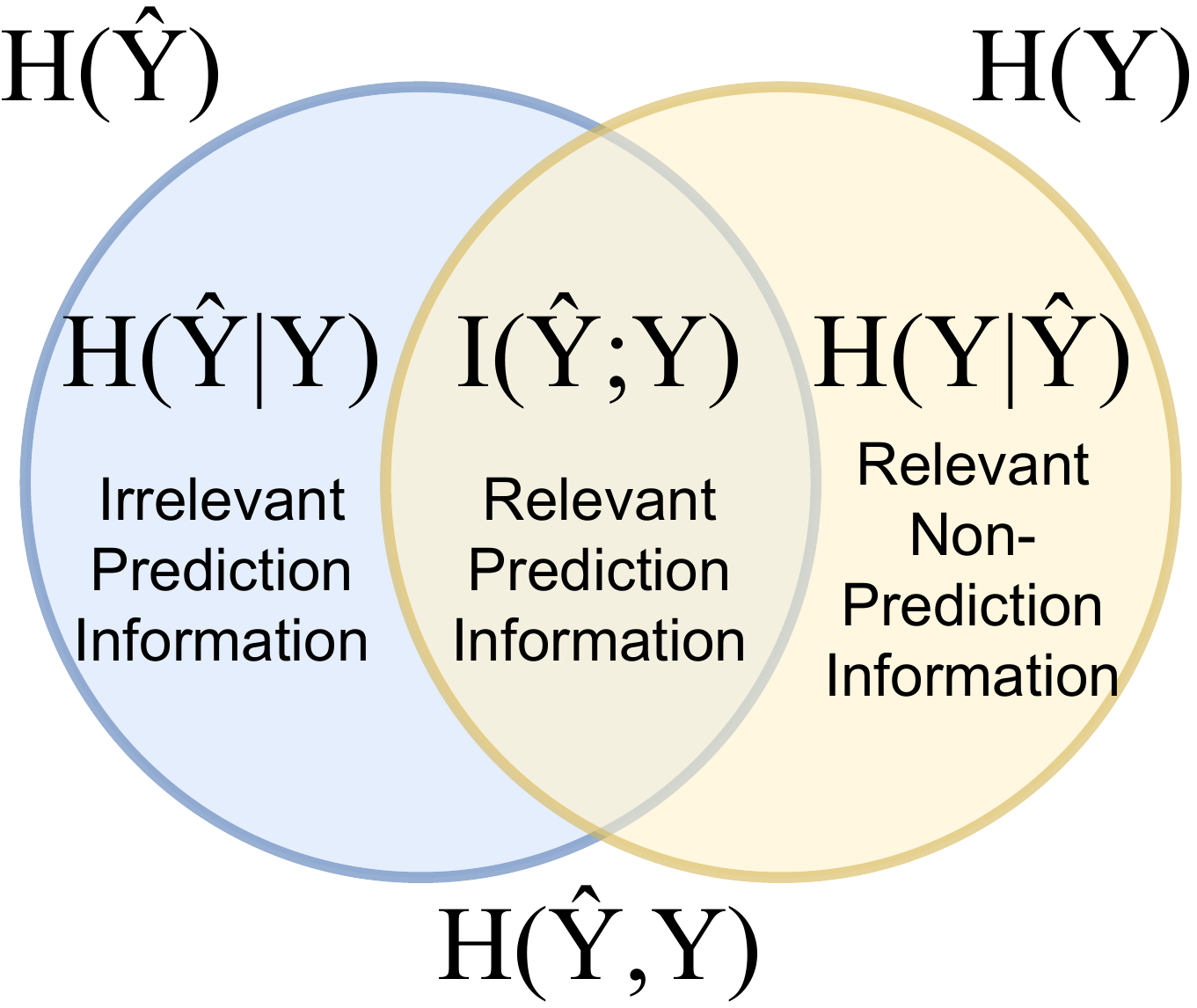}
        \label{fig:uncertainty_venn_yyhat}
    }
    \caption{
        Venn diagrams of the subtractive and additive relationships between information measures of the random variables for input $X$, task label $Y$, and predictions $\hat{Y}$.
        These sets are the types of uncertainty that occur in black box evaluation.
        Epistemic uncertainty in this paper is the conditional entropy $H(\hat{Y} | Y)$.
        Estimating these uncertaintes in practice tends to be intractible.
        The Bayesian Evaluator only needs to sample from $P(\hat{Y}|Y)$ to measure epistemic uncertainty through a credibility interval.
        More details in Sec.
        ~\ref{sec:background}--\ref{sec:Bayes}
        , SM~2.3, SM~3.3, and SM~3.4..
    }
    \label{fig:uncertainty_venn}
\end{figure*}
\textbf{Task Relevant Information.}
The supervised learning task is defined by the input feature data and task labels, respectively represented in this work by the random variables $X$ and $Y$. 
A predictor learns the mapping of the input to the labels, resulting in predictions represented by the random variable $\hat{Y}$.
In information theory, the mutual information $I(X;Y)$ is the information shared by both $X$ and $Y$
~\cite{shannon_mathematical_1948}.
The labels' information defines the relevant information to the task, and their mutual information with $X$ is the relevant information in the input that is desired to be kept in the predictions $\hat{Y}$, as stated in the Information Bottleneck
~\cite{tishby_information_2000}
and seen in Fig.
~\ref{fig:uncertainty_venn}.
The remaining information in $X$ is deemed irrelevant to the task and desired to be ignored in the final predictions~[SM~3.2].
The predictor may include more or less information and even introduce more irrelevant information based on how they learn the task, \textit{e.g.} inductive bias~[SM 3.3].
Uncertainty in the context of the uncertainty estimation literature
~\cite{abdar_review_2021}
is the irrelevant information in this formulation, with the conditional entropy $H(\hat{Y}|Y)$ being the epistemic uncertainty, which is the focus of this work.
This paper refers to irrelevant information as uncertainty to unify these different perspectives and to Shannon-information as information~[SM 2.3].

Using the Maximum Entropy Principle
~\cite{jaynes_probability_2003}
with the theoretical backing of Markov Chain Monte Carlo (MCMC)
~\cite{neal_bayesian_1996}
, our BNN evaluator of $P_{\hat{Y}}(\hat{Y}|Y)$ samples from the correct posterior distribution as long as MCMC convergence occurs.
This allows us to answer the question, ``What values does the predictor confuse with a given label?''.

\begin{figure*}[ht]
    \centering
    \includegraphics[width=1.0\textwidth]{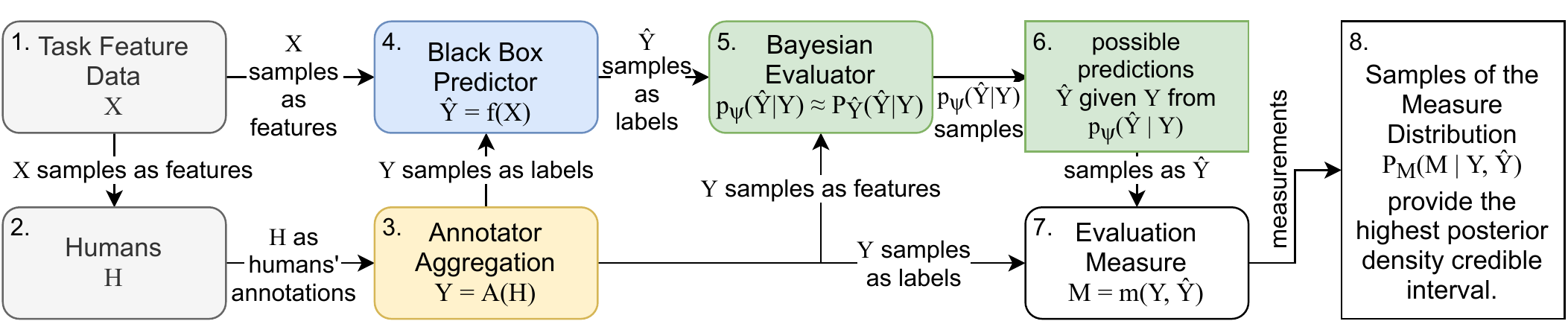}
    \caption{
        The Bayesian evaluation framework depicting the
        Bayesian Evaluator's
        samples standing in for the actual predictions to assess the predictor's uncertainty.
        The human annotations for the feature data are aggregated together, giving each task sample a single accepted label (\textit{e.g.}, human frequency) that is used to fit the supervised learning predictor with the task features.
        An evaluator is chosen to map the accepted labels to the predictor's possible outputs,
        yielding a generative model that approximates the conditional distribution that captures the predictor's uncertainty.
        Any measure may be applied to the annotator aggregations and evaluator's samples, resulting in the measure's distribution.
        The modules are numbered in the typical order performed. See Sec.
        ~\ref{sec:Bayes}.
    }
    \label{fig:framework}
\end{figure*}
\section{A Bayesian Evaluation Framework}
\label{sec:Bayes}

In supervised learning, there are two probability spaces: for feature data $(\Omega_X, \F_X, P_X)$ and for target labels $(\Omega_Y, \F_Y, P_Y)$, where $\Omega$ is the sample space, $\F$ is the event space, and $P$ is the probability measure.
Let $X:\Omega_X \mapsto E_X$ and $Y:\Omega_Y \mapsto E_Y$ be two random variables (r.v.)
, where $(E, \E)$ represents an $E$-valued measurable space with $\sigma$-algrebra $\E$.
Note subscript notation in SM~2.1 with $P_Y$ as the probability measure of $Y$.
$X$ represents the feature data with samples $x \in E_X$ and $Y$ represents the target labels with samples $y \in E_Y$.
The paired data samples of each r.v. are defined as the following $n$-tuple of pairs
$
    \Data = ((x_i, ~ y_i) ~ : ~ i \in \{1, \dots, n\}, x_i \in E_X, y_i \in E_Y)
$
where $n$ is the number of samples, and $(x_i, y_i)$ are the paired samples indexed by $i$.
There is a set of possible functions $F$ that map the input space to the output space $f:E_X \mapsto E_Y$, where $f \in F$.
The goal of supervised learning is to approximate a function $f$ given the data samples $(x_i, y_i)$, resulting in the learned function $f(x)$.
The predictor's output $\hat{y} = f(x)$ serves as samples of its own r.v. $\hat{Y}:\Omega_Y \mapsto E_Y$ with the same sample, event, and measurable spaces as $Y$.

When obtaining crowdsourced annotations, there is a set of humans $\Humans$ where each human $\human \in \Humans$ is a predictor that may be expressed as a r.v. $\human_j(x_i) = h_{i,j}$, where $h_{i,j}$ is the $j$-th human's annotation for the sample $x_i$.
This results in a vector of human annotations $\vec{h}_i$ per sample $x_i$
\begin{equation}
    \vec{h}_i = (h_{i,j} ~ : ~ j \in \{1, \dots, m\})
\end{equation}
for $m$ unique humans.
The multiple vectors of human annotations per sample $x_i$ forms an $n$-tuple $H~=~(\vec{h}_1, \vec{h}_2, \dots, \vec{h}_n)$.
The paired feature data and human annotations $(x_i, \vec{h}_i)$ serves as input to the framework under steps 1 and 2 in Fig.~\ref{fig:framework}.
In this paper's experiments, the feature data as samples of $X$ are images and each human provided a single class as an annotation forming a vector of nominal data for $\vec{h}_i$ .

Given the need for target labels $Y$ in supervised learning, an annotator aggregation $A(\cdot)$ is selected by the designer to define what is the task-relevant information~\cite{tishby_information_2000} contained within $H$ for the task to be learned.
The annotator aggregation is then performed on each $\vec{h}_i \in H$ yielding the samples of the desired r.v. $Y$:
\begin{equation}
    A(H) := (A(\vec{h}_i) ~ : ~ \vec{h}_i \in H, ~ A(\cdot) \in E_Y)
\end{equation}
For this paper's experiments, the distributions over $Y$ and $\hat{Y}$ are both a distribution over probability vectors.
Many machine learning classifiers use the label space of probability vectors either as hard targets with one hot vectors or soft targets with probabilities.
The soft target is a more informative label that results in more robust classifiers and leads to results that better represent probability vectors \cite{hinton_distilling_2015, peterson_human_2019}.
Thus, this work focuses on the frequency of the nominal annotated class labels $\vec{h}_i$ per task sample image $x_i$ as the annotator aggregation, \textit{i.e.} human frequency, which provides a probability vector as the label $y_i$ for each task sample $x_i$. 
In this case, the label space is the $(K - 1)$ probability simplex $\Delta^{K-1}$, where $K$ is the dimension of the probability vector:
\begin{equation}
    \Delta^{K-1} = \{z \in \R^{K - 1} : z_k \in [0, 1] \text{ and } \sum^{K - 1}_{k = 1} z_k \leq 1 \}
\end{equation}
This is the same as the sample space of a Dirichlet distribution~\cite{minka2000estimating}.
The event space is then $E_Y = \Delta^{K-1}$ for labels and $E_X$ is the space of images for feature data.
The elements of the label and prediction sample spaces are thus probability vectors in $\Delta^{K-1}$.
However, these probability vectors are single point estimates of the distribution being modeled, thus our framework does not only estimate uncertainty from a single probability vector~\cite{gal_uncertainty_2016} common in calibration~\cite{widmann2020calibration}, but also from the distribution of probability vectors.
As point-estimate labels, the aleatoric uncertainty from the nominal human annotations is captured no further than this frequency as a probability vector, leaving the examined source of uncertainty to be from the predictor alone in this work's experiments.

With this problem formulation, there are three r.v.s, $X$, $Y$, and $\hat{Y}$, all of which have distributions over them that may be estimated from the available data.
The distribution over r.v.s is known to capture the uncertainty of the phenomena it models and whose properties may be used to assess that uncertainty.
Examples include information entropy~\cite{shannon_mathematical_1948}, which is the expected information of an r.v., and the HPDI~\cite{ohagan_kendalls_2004}.
Given that these distributions are estimated from data in this paper's experiments, an HPDI is used for simplicity instead of estimating the entropy or mutual information to avoid estimation error that occurs due to the curse of dimensionality. 
The following conditional distribution captures a predictor's expected epistemic uncertainty in $\hat{Y}$ with regard to $Y$, also known as the irrelevant information of the predictions
~\cite{tishby_information_2000}
\begin{equation}
    P_{\hat{Y}}(\hat{Y}|Y) := \{P_{\hat{Y}}(\hat{y}|y) ~ : ~ y, \hat{y} \in E_Y\}
    \label{eq:cond}
\end{equation}
where $y$ is a target label sample from $Y$ and $\hat{y}$ is the predictor's output given the respective feature data sample $x$.
This conditional distribution may be approximated from the available data $(y_i, \hat{y}_i)$.
The predictor being trained with the target labels results in a conditional relationship forming between the predictor's output given the target label when the predictor learns the task, thus having the two r.v.s become more similar to one another as the task is learned~\cite{tishby_information_2000}.
Eq.~\ref{eq:cond} only requires the r.v.s $Y$ and $\hat{Y}$, thus it is agnostic of the predictor that produced the samples of $\hat{Y}$.
As long as enough data is available, the source of the labels $Y$ also does not matter, thus making this agnostic to the chosen annotator aggregation.

The conditional distribution in eq.
~\ref{eq:cond}
captures the epistemic uncertainty in the predictor by expressing the possible predictions $\hat{y}$ and their probabilities given the accepted label $y$.
Similar to the supervised learning task in finding a function $f(x) = y$ from all possible functions of $f$, these possible predictions are the transformations $t(y) = \hat{y}$ from all possible transformations $t \in T$ of the label's event set $E_Y$.
To model this conditional distribution and find the possible transformations of $t$ given the data and any prior knowledge, the Bayesian Evaluator in Fig.
~\ref{fig:framework}
is the following Bayesian probablistic model
\begin{equation}
    p_\psi(\hat{y}|y) \approx P_{\hat{Y}}(\hat{y}|y)
\end{equation}
where $\psi$ is the parameter set of the probablistic model.
Fig.~\ref{fig:framework} shows that the inputs to this Bayesian Evalutor are the human frequencies used as its "features", while the predictor's outputs are used as the evaluator's target labels.
The posterior distribution may be approximated via MCMC methods, such as the Hamiltonian Monte Carlo (HMC) algorithm \cite{duane_hybrid_1987}, resulting in a generative model that can be sampled from to perform Bayesian analysis.
If $p_\psi$ samples from the posterior distribution of $P_{\hat{Y}}(\hat{Y}|Y)$, as in MCMC methods~\cite{neal_bayesian_1996}, then the Bayesian Evaluator may be relied upon to evaluate the epistemic uncertainty of the predictor.

While there is variance across the samples of $X$, thus resulting in task-relevant and irrelevant information, the epistemic uncertainty we want to evaluate is that per prediction $\hat{y}$.
The Bayesian model $p_\psi(\hat{y}|y)$ is able to leverage information about the epistemic uncertainty from both data and \textit{a priori} knowledge, under the constraints of the Bayesian evaluator's design choices (\textit{e.g.}, which prior distribution is used to express the \textit{a priori} knowledge).
To focus only on the epistemic uncertainty we used human frequency as the annotator aggregation to obtain point-estimate training labels, which is the mean of nominal values without variance, \textit{i.e.}, aleatoric uncertainty.
This design decision simplifies the presentation of this evaluation framework, which is still capable to be used in such a case.

In the supervised learning context, any information in that $Y$ is task-relevant as defined in the Information Bottleneck
~\cite{tishby_information_2000}, including the label's alreatoric uncertainty.
Handling the uncertainty between annotators and determining what part of the annotator's joint entropy is relevant to the task is ongoing research in annotator aggregation~\cite{zheng_truth_2017, hung2013evaluation}.
Given the focus on epistemic uncertainty, the information on the uncertainty expressed by the model $p_\psi(\hat{y}|y)$ is determined by the available data, the priors, and the model's design.
We followed the maximum entropy principle~\cite{jaynes_probability_2003}, meaning that all priors are assumed uniform letting the data provide most of the information on the uncertainty.

To leverage data to evaluate the epistemic uncertainty, multiple predictions per sample are helpful, but not necessary.
If multiple predictions per sample are available and representative of the predictor, then it would provide more data to learn the possible mappings of $t$.
There are various ways to obtain these predictions per sample, albeit at a great cost of computation time and resources.
If the predictor is purely non-stochastic, then fitting the Bayesian model to instances of the predictor trained on different folds in K-fold cross validation would provide the data that partially expresses the uncertainty.
Of course, a larger number of predictions per sample is more desirable, and in practice K-fold cross validation is done with single digit values for K.
Alternatively, if the predictor uses stochastic training, \textit{e.g.}, Stochastic Gradient Descent, one could train the predictor on the same fold with different random seeds.
For ANNs, options such as drop-out at test time or perturbing the weights may be used to obtain multiple predictions per sample.
Existing calibration and epistemic uncertainty estimation methods that model $P_Y(Y|\hat{Y})$, rather than Eq.
~\ref{eq:cond}
, may be used to provide these multiple predictions as well.
This paper's experiments that showcase the evaluator do not use multiple predictions per sample, relying on the evaluator's design and the data to express the expected epistemic uncertainty.

After the generative model of the posterior distribution has been obtained as the Bayesian Evaluator, samples of $p_\psi(\hat{y}|y)$ may be used along with the task target label $y$ to obtain the conditional distribution over an arbitrary measure $m(y, \hat{y})$ of the predictor:
\begin{equation}
    P_M(M | Y, \hat{Y}) = \{P_M(m | y, \hat{y}) : y, \hat{y} \in E_Y\}, \text{where} ~ M = m(Y, \hat{Y})
\end{equation}
The term measure here is used in the context of measure theory~\cite{tao2011introduction},
where metrics are a subset of measures and cannot add information
~\cite{cover_elements_1991}.
With the conditional distribution over the measure obtained, properties of the measure's distribution may be used to assess uncertainty of the measure~\cite{ evans_probability_2010, shannon_mathematical_1948}, such as the HPDI~\cite{ohagan_kendalls_2004}.
The smaller the HPDI, the better because then there is less spread of possible values of the measure for the given credibility score.
This aids in predictor selection to help not only find the best performing predictor, but also the least uncertain one.
Fig.~\ref{fig:framework} depicts the framework in its entirety.
In contrast to calibration~\cite{guo_calibration_2017, abdar_review_2021}, this framework is for evaluating the epistemic uncertainty of the predictor and does so without modifying the predictor itself.
In this work, two evaluators are implemented showcasing the proposed framework, but \textit{any sufficiently strong} Bayesian model $p_\psi(\hat{y}|y)$ may be used as the evaluator.
Both rely on a transformation of the label space in SM~3.6.1.

    \textbf{Normal Distribution Over the Differences.}
    A simple statistical evaluator is implemented that models the distribution over the differences $y - \hat{y}$ within the probability simplex as a multivariate normal
\begin{equation}
        p_\psi(\hat{y}|y) \sim y + \N(\mu, \Sigma)
    \label{eq:dod_mvn}
    \end{equation}
    where $\psi$ consists of $\mu$ as the means and $\Sigma$ as the covariance matrix of the differences.
    This evaluator will be referred to as the Normal Distribution over the Differences (NDoD).
    Given the difficulty in modeling a conditional distribution within the probability simplex, resampling is used to avoid output values that are not proper probability vectors.
    Resampling in high dimensions results in extremely low probability that the sample is within the probability simplex. Thus a downside to this approach is that it is relatively inefficient computationally.
    The parameters of the normal distribution are determined analytically via uniformly minimum-variance unbiased estimators of the mean and covariance of the differences.
    This approach yields the desired distribution without the need of HMC albeit with more limiting assumptions about the possible distributions of difference.
    Given that NDoD models the differences overall, irrespective of the individual target labels, intuition leads to the unbiased estimation of the differences' mean to be zero for predictors that perform well in their predictions.
    This means that the random function $t(y)$ is expected to be the identity function with some Gaussian noise when using NDoD.
    Given this, we also examine the Zero Mean NDoD.

\textbf{Bayesian Neural Network}
The second evaluator is more general, and makes minimal assumptions about the phenomena characterized by the data in accordance with the maximum entropy principle \cite{jaynes_probability_2003}.
Following this principle allows the data to have a maximum effect on the resulting posterior distribution given a set of model design decisions, which is desired when no prior knowledge is available.
Being independent of prior domain knowledge allows the evaluator to be agnostic to the predictor and target labels.
Thus we use a BNN
~\cite{neal_bayesian_1996}
fit using HMC
~\cite{duane_hybrid_1987}, which handles the potential for non-linear transformations in $t(y) = \hat{y}$, with a uniform prior distribution over the weights and biases $\psi$.
The BNN learns the possible transformations at different points in the probability simplex $\Delta^{K-1}$.
The BNN's inputs are the target labels $y$ in the $K-1$ simplex space post-transformation.
The BNN outputs its approximation of the predictor's output $\hat{y}$ in the same space.
The output is then transformed back into the original $K$ dimensional space and mapped through the softmax function to stay within the probability simplex's bounds and to avoid costly resampling.
For BNN implementation details, see SM~3.6.2.
The BNN's target likelihood function is that of a multivariate normal about the differences between the output of the BNN and the predictor's output $\hat{y}$
\begin{equation}
    \Like(\psi|y, \hat{y}, \sigma) = \N(\text{BNN}(y, \psi) - \hat{y}, ~ \sigma^{2}\text{I})
\end{equation}
where $\sigma$ is a hyperparameter that stands for the standard deviation, I is the $K \times K$ identity matrix, and their product serves as the covariance of the multivariate normal.
The standard deviation hyperparameter in this case determines the lower bound of the variance of the BNN's output per predictor's output.
As the lower bound, the standard deviation can be used to increase or decrease the lower bound of the evaluator's variance.
The HMC for each dataset converged on average in 3 days~[SM 4.2.1].

\section{Experiments}
    Four computer vision datasets for recognition tasks involving subjective judgements were used: SCUT-FBP5500~\cite{liang_scut-fbp5500:_2018} for assessing human opinions on facial beauty, social attribute assignment of trustworthiness using data from TestMyBrain.org~\cite{germine2012web}, apparent age estimation using the APPA-REAL~\cite{agustsson2017appareal} data from the ChaLearn series of challenges, and LabelMe~\cite{rodrigues_learning_2017} for ambiguous scene classification.
    To obtain a probability vector as the label space of the task, the human frequency as annotator aggregation was obtained by treating all individual human labels as discrete class labels.
    Data in LabelMe, SCUT-FBP5500, and TestMyBrain.org data were already discretized.
    The apparent age labels in APPA-REAL were discretized from the original range of 0 to 100 into bins spanning 5 years. 
    The resulting number of classes was 8 for LabelMe, 5 for SCUT-FBP5550, 7 for the TestMyBrain.org data, and 20 for APPA-REAL.
    More in  SM~4.1.
    
\begin{figure*}[ht]
    \centering
    \includegraphics[width=\textwidth]{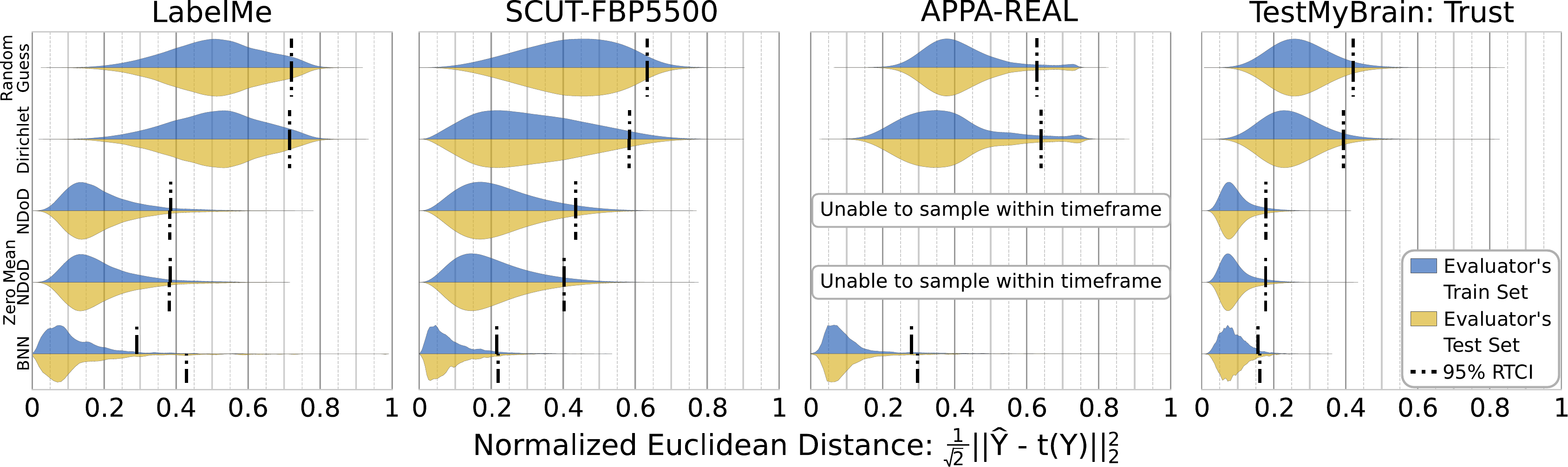}
    \caption{
        Experiment 1 shows how well each Bayesian Evaluator fits the conditional distribution $P_{\hat{Y}}(\hat{Y} | Y)$ of the predictor output given the human frequency by measuring the distances between the models' samples to the predictor's output.
        Higher density near zero indicates a better fit model of $P_{\hat{Y}}(\hat{Y} | Y)$.
        95\% of the probability density is to the left of the right-tailed credible interval showing the models' performances are quantitatively different.
        All models are fit on half of the combination of the predictor's training and test sets, and are generalized as they perform similarly on data they were not fit on.
        NDoD and the BNN capture the conditional relationship better than the evaluators that $P_{\hat{Y}}(\hat{Y}|Y) = P_{\hat{Y}}(Y)$.
        The BNN best matches the predictor's output.
    }
    \label{fig:exp1_train}
\end{figure*}

\begin{figure*}[t]
    \centering
    \includegraphics[width=\textwidth]{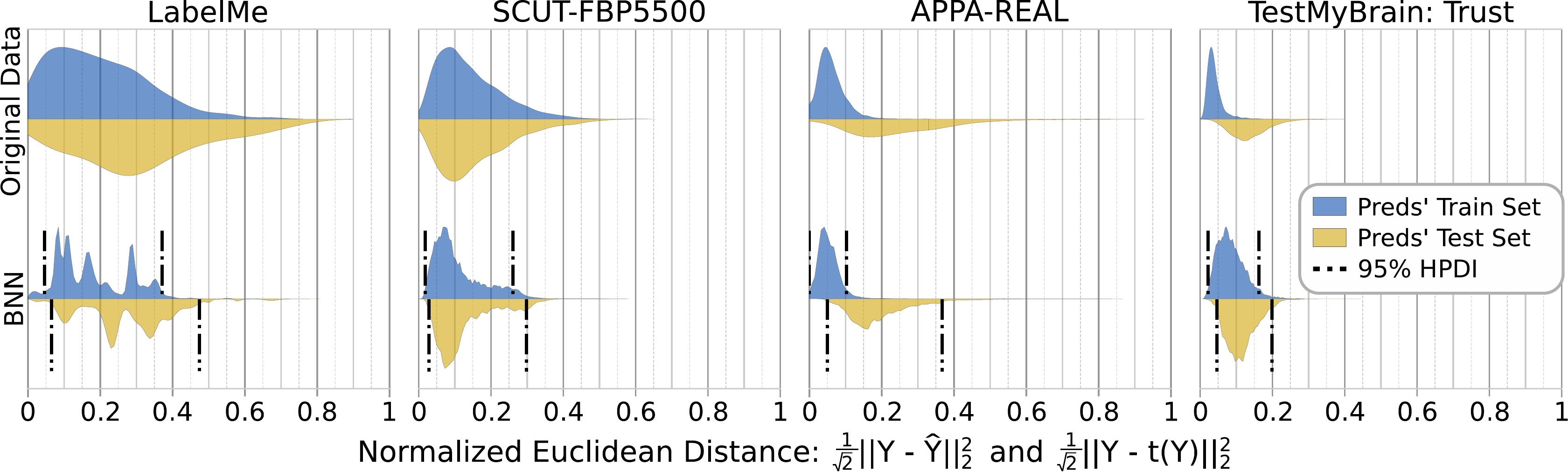}
    \caption{
        Experiment 2's results show the $L_2$ distribution and 95\% HPDI obtained from a BNN model compared to the original data and how it may be used in practice.
        The model consists of two BNNs fit and applied separately on the training and test sets in case their distributions differ.
        Given the data, model, and priors, the measure distribution indicates the expected Euclidean distance of the predictor's output to the target human frequency.
        The HPDI is able to be obtained with our framework, but not without.
        The 95\% HPDI indicates better performance and certainty in the training sets than in the test sets due to the higher density in lower values.
        The TestMyBrain Trust predictor is best performing and certain in both training and test sets.
    }
    \label{fig:exp2}
\end{figure*}

\begin{figure*}[t]
        \centering
        \includegraphics[width=\textwidth]{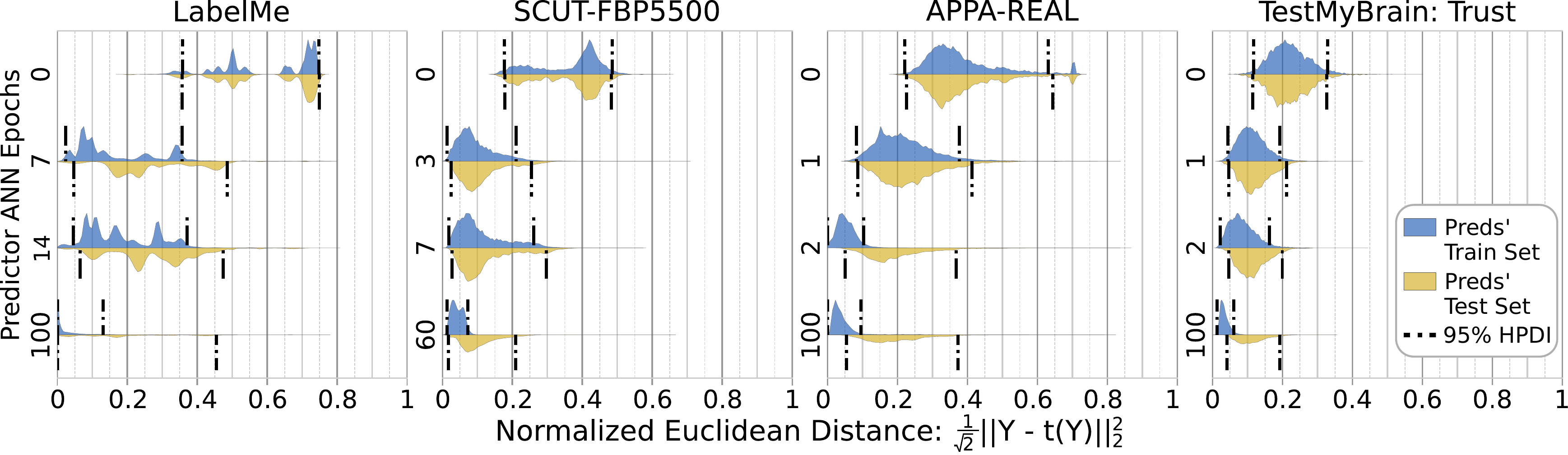}
        \caption{
            Experiment 3 shows the use of the framework in practice to aid in model selection of a predictor trained over epochs.
            The best performing and least uncertain model fit is identified by assessing the epistemic uncertainty through the use of the 95\% HPDI of the Euclidean distances.
            The ideal predictor has the highest density in the smallest distances, meaning it is the closest to the target labels, and has the smallest HPDI indicating it is the least uncertain.
            The preference for performance to certainty depends on the application at hand.
            The framework captures the expected tendencies of training over time, providing informative feedback and a quantitative comparison of credible intervals.
        }
        \label{fig:exp3}
    \end{figure*}

    All predictors were trained by combining and then splitting the available annotator data into a single pair of partitions with 80\% training samples and 20\% test samples.
    A deep neural network was trained for each of the four datasets to serve as the predictor.
    Each network was trained in a manner that was consistent with established practices for the datasets [SM~4.2]. 
    After training the predictors and obtaining their predictions for their respective training and test sets, the two Bayesian Evaluators are applied and examined.
    For measures, we examined Normalized Euclidean Distance ($L_2$), KL divergence, and AUC [SM~4.3].
    The following experiments focus on  $L_2$, which is the distance of one point to the target point.
    In Figs.~\ref{fig:exp1_train}--\ref{fig:exp3}, lower values are more accurate, with smaller credible intervals being more precise.

    \textbf{Experiment 1: Bayesian Evaluator Assessment.}
    In order to determine if NDoD and the BNN are capable of modeling the conditional relationship between $Y$ and $\hat{Y}$, they were compared against two baseline distributions that assumed independence between $Y$ and $\hat{Y}$, \textit{i.e.}, $P_{\hat{Y}}(\hat{Y}|Y) = P_{\hat{Y}}(\hat{Y})$.
    These two models were ``random guess'' implemented as a uniform Dirichlet distribution and a maximum likely Dirichlet distribution using the alternative parameterization of mean and precision~\cite{minka2000estimating} [SM~4.4].
    To assess the Bayesian Evaluators' fits and generalization, each predictor's train and test sets were combined, shuffled, and then split in half (avoiding overfitting partitions) to create the train and test sets for the Bayesian Evaluators of $p_{\psi}(\hat{y}|y)$.
    This experiment assesses only the Bayesian Evaluators, where different evalutors are compared to one another by how well they represent the predictor's outputs, thus shuffling the data does not impact the original task performance.
    Results for this experiment are found in Fig.~\ref{fig:exp1_train} and SM~4.4.
    
    The $L_2$s between the predictor's output $\hat{y}$ and the Bayesian Evaluator's outputs were used to indicate how close the different models of the conditional distribution are from the actual data.
    The desired properties of the Bayesian evalators are short $L_2$ and minimal spread, indicating output that consistently matches the real data.
    A right-tailed credible interval (RTCI) was used to compare the models' fit of $P_{\hat{Y}}(\hat{Y}|Y)$.
    The RTCI is the interval over the measure where there is 95\% of the probability density to the left of the interval line, leaving 5\% to the right of the interval line~[SM~2.2.2].
    The evaluator with the lower-valued interval line, thus the smallest RTCI, is the best fitting model.
    
    Based on their RTCI, higher densities in shorter $L_2$, NDoD and the BNN both successfully model the conditional relationship between $Y$ and $\hat{Y}$, relative to the baselines that assumed independence.
    The zero mean NDoD was the better of the two NDoD variants.
    However, the BNN either matched or had less distance from the actual predictions, as seen by its consistently lower $L_2$ measurements than either NDoD implementation.
    Notably, both NDoD variants were unable to finish sampling the conditional distribution for APPA-REAL due to resampling in high dimensions [SM~4.4].
    The BNN's performance is somewhat questionable only when compared to NDoD for LabelMe.
    However, given that most of the 95\% density of the BNN's $L_2$s to the predictor's output is denser near zero than both NDoD models, and that the mean and medians also indicate this trend, the BNN is still a quantitatively better model of the conditional distribution than NDoD for LabelMe.
    This experiment was also run over multiple folds [SM~4.4.2].
    Given the BNN being the best evaluator, it is used in experiments 2 and 3.

    \textbf{Experiment 2: Predictor Uncertainty Assessment.} 
    This experiment reflects how the full framework would be used in practice to obtain a HPDI for the performance measure of a single predictor.
    The BNN model is deployed to assess the empirical uncertainty of the measures by being used in place of the predictor's actual predictions.
    To handle possibly different conditional distributions in the training and test sets, a BNN is fit to each separately and is used to assess its respective set.
    Sampling from the BNN and calculating the measure between its output given the target labels provides the distribution of measures, which inherently contains the uncertainty of the measure~\cite{shannon_mathematical_1948, jaynes_probability_2003} as experessed in the 95\% HPDI.
    In Fig.~\ref{fig:exp2}, the original data's $L_2$s from the target label to the predictor is compared to the distance of the target label to the BNN's output given the human frequencies.
    The HPDI can be obtained given the model and its priors, where the original measurements alone are unable to express their credibility.
    
    The BNN's HDPI indicates those measures occurring in that interval 95\% of the time.
    The BNN for LabelMe has the most different measure distribution, which can be interpreted as LabelMe's inherent uncertainty caused from both the task's difficulty and an uneven distribution of human annotations [SM~4.1].
    All of this, in addition to having the largest 95\% HPDI, indicates that the LabelMe predictor is the least certain predictor with the worst performance.
    Based on these results, the predictors for TestMyBrain and APPA-REAL are very certain predictors for their training sets because their $L_2$ distributions have the smallest 95\% HPDI.
    TestMyBrain's predictor is the most certain on a test set.
    SCUT-FBP5500's predictor is the most consistent in both certainty and performance on the train and test sets.
    Results for KL divergence and AUC are in SM~4.5.1.

\textbf{Experiment 3: Predictor Certainty Over Training Periods.}
    The framework was also used to assess the uncertainty of underfit and overfit predictors to depict the use of the conditional distribution in model selection.
    The data partitions and the BNN usage are the same as in Experiment 2.
    First, each deep neural network predictor was randomly initialized.
    Second, a number of epochs was chosen based on the mean $L_2$ of the Experiment 2 predictions to the human frequency.
    From this, the halfway epoch to finished fitting was selected.
    Third, a larger number of epochs was selected via the different termination methods of the predictors, such as early stopping for LabelMe and SCUT-FBP5500.
    And finally, a number of epochs well past a predictor's recommended termination was chosen, which was intended to represent an overfit predictor.
    
    In Fig.~\ref{fig:exp3}, there is a trend for all of the datasets:  as the predictor continues to train, its uncertainty in the conditional distribution decreases for the training set.
    This is visible in the decrease in width of the 95\% HPDI.
    This follows suit with the understanding that the deep neural networks are converging to various minima and their results become more consistent for the given task.
    This trend also occurs in the test set, albeit to a much lesser extent.
    Once overfitting begins, the variance may increase again, as seen in the LabelMe's predictor's HPDI length increase for the test set.
    Also, with the decrease in overall variance, there is  noticeable training improvement based on the density of the $L_2$s.
    The $L_2$s not only get tighter in density with a smaller 95\% HPDI, but also tend to approach zero, indicating the predictor's improvement on the data. 
    Thus, the BNN is capable of reflecting the uncertainty associated with an overfit predictor.
    With this ability to assess the uncertainty of predictors, model selection can be improved by finding the most certain best performing model.

\section{Conclusion}

In summary, we have introduced an evaluation framework for analyzing the epistemic uncertainty of predictors trained with subjective annotations using Bayesian modeling of the conditional distribution $P_{\hat{Y}}(\hat{Y} | Y)$.
Two implementations demonstrate the framework and the more general model, the BNN, was capable of capturing conditional relationships across four different datasets for  visual recognition problems involving subjective human judgement.
The framework provides a modular approach to estimating the uncertainty of predictors, remaining agnostic to the classifier, data, and task domain.
Further developments in more efficient and transparent BNNs or other Bayesian models for use in this framework are of interest for future work, as well as modeling the aleatoric uncertainty in Bayesian Inference.
The broader impacts of this work are included in SM~5.

\section*{Acknowledgements}
We thank Timothy J. Ireland, Jeffery Kinnison, Daniel Gonzalez, and Matthew Kaufman who provided valuable feedback on drafts of this paper.
Also, we thank Timothy J. Ireland for the lengthy discussions on the theory of this work and related topics.

\bibliographystyle{elsarticle-num} 
\bibliography{bibs/data_background.bib, bibs/framework.bib}

\newpage
\setcounter{section}{0}
\renewcommand*{\theHsection}{SM.\the\value{section}}
\section{Supplementary Material}

This document is the Supplementary Material (SM) of the main text for ``\papertitle''.
The content of this document is intended to provide a convenient refresher on background topics pertaining to the Bayesian Evaluation Framework in SM~Sec.~\ref{sec:bg}, extra details on the nuance of the framework in SM~Sec.~\ref{sec:bef}, and the specific implementation details for reproducibility of the experiments conducted in the main text SM~Sec.~\ref{sec:sm_exps}.

\section{Supplementary Material for Relevant Background Topics}
\label{sec:bg}

This section contains helpful background knowledge for the convenience of the reader to aid in understanding the Bayesian Evaluation Framework, in case a refresher on some topics is necessary.

\subsection{Notation and Keywords}

The main text relies on standard notation for probability and statistics from the Bayesian Inference perspective, as well as Shannon-information theoretic measures~\cite{shannon_mathematical_1948}.
The notation used in the main text and this SM is listed in Table~\ref{tab:notation}.
This notation follows from the measure theoretic standard for probability theory, specifically for random variables.
The notation for a probabilistic model, \textit{e.g.}, $p_\psi$, is standard in Variational Bayesian Inference where the probabilities returned by $p_\psi$ depends upon the specific model instantiation that uses the parameter set $\psi$. Notable keywords are as follows:
\begin{itemize}
    \item \textbf{Supervised Learning Task}: 
        Given the input feature data $x \in E_X$ and their paired target labels $y \in E_Y$, learn a mapping $f:E_X \mapsto E_Y$ from all possible mappings $f \in F$.
        Defined in main text Sec.~3
    \item \textbf{Predictor}: 
        The algorithm or process used to learn or perform the supervised learning task.
        Defined in main text Sec.~3.
    \item \textbf{Subjectively Annotated Task}: 
        A Supervised Learning Task where objective ground truth is not available to serve as the target labels, but subjective labels are accepted as the target label.
        The target labels are obtained from humans who provide annotations as their answer to the task where you can see each human as a different predictor for the task.
        Defined in main text Sec.~1 at a high level with the process detailed in Sec.~3.
    \item \textbf{Annotator Aggregation}: 
        The function that given human annotations for a single feature data sample $x$ creates a label that represents the task-relevant information.
        Defined in main text Sec.~1 and Sec.~2 at a high level with the process detailed in Sec.~3.
    \item \textbf{Bayesian Evaluator}: 
        The algorithm or process used to evaluate a predictor's possible predictions and their probabilities given a target label, which indicates what values a predictor confuses with the target label and to what degree of confusion.
    \item \textbf{Measure}: 
        A measure as defined in measure theory~\cite{tao2011introduction}.
        Mentioned in main text Sec.~1--3 and SM~3.1.
        Note that the measure, as a function, cannot add information meaning it can only maintain or lose the information of its inputs~\cite{cover_elements_1991}.
    \item \textbf{Measure Distribution}: 
        The probability distribution of the measure when given random variables as its input.
        The measure distribution is what 95\% credible intervals are applied over to find the measurements that occur 95\% of the time.
        Defined in main text Sec.~3.
\end{itemize}

{
\rowcolors{1}{white}{light-gray}
\small
    \begin{longtable}{p{0.32\textwidth}p{0.63\textwidth}}
        \caption{
            The notation and definitions used throughout the main text and this supplementary material.
            For further background information on these concepts please refer to their related references.
            Most probability concepts are covered in~\cite{jaynes_probability_2003, ohagan_kendalls_2004}, information theoretic measures in~\cite{shannon_mathematical_1948}, measure theory in~\cite{tao2011introduction}, and Bayesian Inference and modeling in~\cite{neal_bayesian_1996}.
            \label{tab:notation}
        } \\
        Notation & Definition \\
        \hline
        \endfirsthead
        Notation & Definition \\
        \hline
        \endhead
        $\Omega$ & A sample space $\Omega$; a non-empty set. \\
        $\F \subseteq 2^\Omega$ & An event space as a $\sigma$-algrebra $\F$ that is a set of subsets of $\Omega$, which are called events. \\
        $P : \F \mapsto [0,1]$ & A probability measure $P$ that maps events to their respective probability.
        Probability density functions and probability mass functions are common examples of a probability measure.
        \\ 
        
        $(\Omega, \F, P)$ & A probability space triple of sample space $\Omega$, event space $\F$, and probability measure $P$. \\
        $(E, \E)$ &
            A measurable space consisting of the paired event set $E$ and $\sigma$-algebra $\E$.
            \\
        \hline
        $R:\Omega_R \mapsto E_R$ &
            The random variable (r.v.) $R$ is a measurable function from a set of possible outcomes $\Omega_R$ to an $E_R$-valued measurable space.
            Capital letters are used to denote random variables.
            Probability triples corresponding to a random variable $R$ are denoted as $(\Omega_R, \F_R, P_R)$.
            Measureable spaces as $(E_R, \E_R)$.
            
            \\
        $r \in E_R$
        & Lowercase letters denote a single event instance of their corresponding random variable.
        \\
        $(\Omega_R, \F_R, P_R)$, $(E_R, \E_R)$ &
            The probability space and its elements associated with random variable R, along with the corresponding measurable space.
            \\
        $P_R(R) :=$ \newline $\{P_R(r) : r \in E_R\}$ &
        The probability measure applied to a single sample from a random variable $r$ is the probability measure of that single event.
        The probability measure of the r.v. $R$ is the probability measure representing the distribution over that random variable $R$, \textit{e.g.} the probability density function or probability mass function.
            \\
        \hline
        $X:\Omega_X \mapsto E_X$, \newline $x \in X := \{X(w) : w \in \Omega_X\}$ & The random variable $X$ of the input to a predictor with input samples denoted as lowercase $x$. \\
        $Y:\Omega_Y \mapsto E_Y$, \newline $y \in Y := \{Y(v) : v \in \Omega_Y\}$& The random variable $Y$ of accepted labels for a supervised learning task with their label samples denoted as lowercase $y$. \\
        $\Data =$\newline$((x_i, y_i)|i \in \{1, \dots, n\})$  &
            A supervised learning data set as a tuple of input $x$ and target label $y$ sample pairs where $n$ is the number of samples indexed by $i$. \\
        $f:E_X \mapsto E_Y, f \in F$ & The resulting function of a predictor that performs the supervised learning task from a set of all possible functions. \\
        
        $f(X) := \{f(x)|x \in X\}$ \newline $\hat{Y} = f(X)$  & The resulting random variable of the predictor's predictions given the input data. Shares the same sample, event, and measurable spaces with $Y$, $\hat{Y}:\Omega_Y \mapsto E_Y$. \\
        $t(\hat{y}) = y, t \in T$ &
            The possible transformations of an accepted label to the predictions of a predictor.
            These set of transformations are those that the Bayesian Evaluator selects from and determines with what probability the predictor may select from them for each accepted label.
            \\
        
        \hline
        $\human \in \Humans$ & The set of human annotators where each human is a random variable of an annotation given an input sample. \\
        $A(H) := \{A(\vec{h})|\vec{h} \in H\}$ & An annotator aggregation function, \textit{e.g.} human frequency. \\
        $m(y, \hat{y})$ & A measure from an accepted label $y$ to a prediction $\hat{y}$. \\
        \hline
        $P_{Y}(Y=\hat{Y}|X)$ \newline
        $= P_{Y}(Y=\hat{Y}|f(X))$ \newline
        $= P_{Y}(Y=\hat{Y}|\hat{Y})$
        & The conditional probability that is modeled in calibration and typical epistemic uncertainty estimation methods. \\
        $P_{\hat{Y}}(\hat{Y}|Y)$ & The conditional probability that is modeled in our Bayesian Evaluation Framework by the Bayesian Evaluator. \\
        $p_\psi(\hat{y} | y) \sim P_{\hat{Y}}(\hat{Y}|Y)$ & The Bayesian probablistic model of the evaluator with parameter set $\psi$ that approximates $P_{\hat{Y}}(\hat{Y}|Y)$. \\
        $p_\theta(y=\hat{y} | x) \sim$ \newline $P_{Y}(Y=\hat{Y}|X)$ & The Bayesian probablistic model of the predictor with parameter set $\theta$ that approximates $P_{Y}(Y=\hat{Y}|X)$. \\
        $H(R)$ \newline 
        $H(R|Q)$,\newline
        $H(R,Q)$
        & Shannon-information Entropy, conditional entropy, and joint entropy. Entropy is the expected information of a random variable.  Conditional information is the information in r.v. $R$ not contained within r.v. $Q$. Joint entropy is the union of information between $R$ and $Q$. \\
        $I(R;Q)$ 
        & The mutual information of two random variables X and Y. 
            When applied to feature data's r.v. $X$ and label r.v. $Y$, this determines the relevant information that is able to be learned  by the predictor.
    \end{longtable}
}

\subsection{Credible Interval}
The credible interval is an interval in which a parameter value will occur with a particular probability, \textit{e.g.}, 95\%, based on the posterior probability distribution of that parameter~\cite{ohagan_kendalls_2004}.
Different types of the credible interval may be used based on the desired application, while the highest posterior density interval is often preferred.
The credible interval is similar to the frequentist confidence interval, but the nuance in their definition differs.

\subsubsection{Highest Posterior Density Interval (HPDI)}
The 95\% highest posterior density interval is the interval that has the tightest possible spread of 95\% density of the distribution.
This is accomplished by selecting the moving window of 95\% density across the distribution with the smallest difference between the beginning and end of the interval.
With the desired density interval always remaining constant, 95\% in this case, the smallest such interval will always contain the largest density region of the distribution.
The HPDI typically is the preferred credible interval as it will contain the median of the distribution in which it is applied over, otherwise known as the maximum \textit{a posterori} (MAP)~\cite{ohagan_kendalls_2004}.
The HPDI is also known to be proven to be the best confidence interval for any location or scale parameter~\cite{jaynes_probability_2003}.

\subsubsection{Right-Tailed Credible Interval (RTCI)}
The RTCI is the Bayesian Credible interval equivalent to the right-tailed confidence interval. 
The reason one uses such an direction oriented tail interval is because they only care about checking one direction of the distribution the interval is over.

For example, in the main text's Experiment 1, we used the 95\% RTCI because we want to compare which distribution has the 95\% of their density in the lowest values of the normalized Euclidean distance.
Selecting the Bayesian Evaluator whose RTCI was at the lowest value means that evalutor has 95\% of its distances closer to zero than any of the other potential evaluators.
The HPDI could have been used instead of the RTCI to assess which evaluator not only has the lowest maxium of the inteval, but also the lowest minimum of the interval.
With that information, the designer could then decide how that spread relates to the decision making process of evaluator selection in addition to which predictor's interval is the tightest, and thus most consistent.

\subsection{Information Theoretic Measures}
\label{sec:sm:info}

Entropy is the expected information of a random variable~\cite{shannon_mathematical_1948}.
For clarity in the main text and to avoid undesirable misinterpretations given multiple referenced fields using ``uncertainty'' in different ways, we refer to the irrelevant information as uncertainty.
This follows the common use of ``uncertainty'' in pattern recognition and uncertainty estimation literature as error or undesirable variance.
As noted in the main text's Fig.~1, the $H(\hat{Y}|Y)$ is the irrelevant information that corresponds to the epistemic uncertainty.

 \section{Supplemental Material for the Bayesian Evaluation Framework}
\label{sec:bef}
This section provides the supplementary material for the methodology of the Bayesian Evaluation Framework (Sec.~3 of the main text).
This supplemental material is intended to provide further detail for the specifics of the implementations of the framework to aid in reproducibility of the work.
For clarity, ``SM" is used to indicate content that is contained within this supplementary material document.
References to the main text are stated as such, as are references to the other appendix sections.
    
\subsection{A General Note Regarding the Framework}
    This framework, regardless of the Bayesian model used as the evaluator, allows for the use of pre-existing measures (\textit{e.g.}, Euclidean distance used as a measure of how far off a prediction is from its respective target label), while adding an assessment of their uncertainty.
    Measure is used in a measure theoretic sense~\cite{tao2011introduction} and as a function, does cannot add any information~\cite{cover_elements_1991}.
    With any measure able to be used, the predictor may be evaluated as normal, and in addition, may have its uncertainty evaluated through the use of this framework.
    By using the 95\% highest density credible interval of the measure, the output for the predictor has an associated score of probability (95\%) that the value occurs within the interval.
    The smaller the interval, the better because then there is less spread of possible values of the measure with the given credibility score.
    This aids in model selection to help not only find the best performing model, but also the least uncertain model.
    The main text's Fig.~2 depicts the framework in its entirety, resulting in the samples of both the conditional distribution over the predictor and the measure.
    In this work, two models are implemented showcasing the proposed framework, but \textit{any} Bayesian model may be used.

\subsection{An Assumption of Stationary Distributions}
    This evaluation framework as written performs its evaluation where the r.v.s $X$, $Y$, and $\hat{Y}$ are all observed at a single time-step where the distributions over these r.vs are assumed to be stationary.
    In some problems, especially time series problems, the distribution may not be stationary and change over time, such as in a stochastic process.
    The evaluation defined in this paper and performed on such problems would be for a single time-step of such scenarios, where the data samples available are what is representative of the r.vs at each time step.
    
\subsection{Gaining Relevant Information External to Input is Impossible}
\label{sec:sm:info_conserve}
As noted in the main text's Fig.~1, it is impossible to gain relevant information not contained in $X$ from the predictor when the predictor has no other random variables that are correlated to the target labels' r.v. $Y$.
This is a conservation of information where a function of an r.v. either maintains or loses information from that r.v., but never adds any information\cite{cover_elements_1991}.
And if there are r.v.s in the predictor, but they are independent to the target label, as in BNNs with converged MCMC or variational Bayesian methods, these r.v.s do not contain any relevant information with the target label $Y$~\cite{cover_elements_1991}.
Gaining irrelevant information from these r.v.s independent of $Y$ is possible as that information is independent of $Y$ and is similar to keeping irrelevant information from $X$ as a source.
The only way to gain that missing relevant information not contained within the current feature data is to add a different feature data source or some other r.v. that includes mutual information with  $Y$ not contained in $X$.
    
\subsection{The Relationship Between $P_Y(y=\hat{y}|x)$ and $P_{\hat{Y}}(\hat{Y}|Y)$}
\label{sec:sm:conds}
Using the notation from SM~Table~\ref{tab:notation}, let there be the three r.v.s involved in a supervised learning problem $X$, $Y$ and $\hat{Y}$, with a learned function that attempts to perform the mapping $f(x) = y$ and yields the sample predictions of $\hat{Y}$, as stated in the main text's Sec.~3.
For this derivation, let the data samples of $X$ and $Y$ be fully representative of the population input data and target labels such that all relevant information is contained within this sampling.
This means the frequentest probability distribution of the target labels not only corresponds to the actual population probability distribution, but also that there are enough samples of $X$ and $Y$ that the mutual information $I(X;Y)$ is also fully represented.
As seen in literature~\cite{abdar_review_2021}, the conditional probability distribution that calibration and typical epistemic uncertainty estimators are both concerned with is 
\begin{equation} 
\begin{aligned}
    P_Y(y=\hat{y}|x)
    & = P_Y(y=\hat{y}|f(x)) \\
    & = P_Y(y=\hat{y}|\hat{y})
\end{aligned}
\end{equation}
Where both calibration and epistemic uncertainty are concerned more with the expected error induced by the predictor for a specific sample pair $(x, y)$.
Given the probability measurements of the possible values for $y$ and $\hat{y}$, the conditional probability measure is obtained
\begin{equation} 
    P_Y(Y=\hat{Y}|\hat{Y})
    := \{P_Y(y=\hat{y}|\hat{y}) : y,\hat{y} \in E_Y\}
    \label{eq:sm:cali}
\end{equation}
The conditional distribution in Eq.~\ref{eq:sm:cali}  is then a subset of the same distribution when not enforcing $Y$ to equal values from $\hat{Y}$.
\begin{equation} 
    P_Y(Y=\hat{Y}|\hat{Y})
    \subseteq 
    P_Y(Y|\hat{Y})
    \label{eq:sm:cali_superset}
\end{equation}
Eq.~\ref{eq:sm:cali_superset} clearly shows the conditional distribution of interest of calibration and epistemic uncertainty is a subset of $P(Y|\hat{Y})$, while our Bayesian Evaluator is concerned with $P(\hat{Y}|Y)$.
$P(Y|\hat{Y})$ being the probability of the actual target labels given the predictions, and $P(\hat{Y}|Y)$ being the probability of the predictions given the actual target labels.
Both respectively stated in simpler terms: ``Probability of the predictor being correct" versus ``The probability of possible predictions given the truth".
This indicates the difference in intended use of calibration for predicting the probability of being correct given prior training, and our Bayesian Evaluation framework of understanding what the predictor confuses with the target labels.

The conditional entropy $H(\hat{Y}|Y)$, which we determined is the epistemic uncertainty based on the Information Bottleneck, is calculated through the use of the joint probability distribution of $Y$ and $\hat{Y}$ and the probability distribution $P_Y(Y)$.
The joint probability distribution may be defined using either conditional distribution
\begin{equation}
\begin{aligned}
    P_{Y,\hat{Y}}(Y, \hat{Y}) & =
    P_Y(Y|\hat{Y}) P_{\hat{Y}}(\hat{Y}) \\
   & = P_{\hat{Y}}(\hat{Y}|Y)P_Y(Y)
\end{aligned}
\end{equation}
Which is then used in the  conditional entropy for the discrete case as
\begin{equation}
    H(\hat{Y}|Y) = - \sum_{y, \hat{y} \in E_Y}  
        P_{Y,\hat{Y}}(y, \hat{y}) \log\frac{P_{Y,\hat{Y}}(y, \hat{y})}{P_Y(y)}
\end{equation}
Thus, either conditional distribution may be used to estimate $H(\hat{Y}|Y)$, although most cases tend to stop at the conditional distribution they care about due to entropy being difficult to estimate from samples in practice, especially in the multivariate continuous cases.
Also, a probability is nicely normalized between $[0, 1]$, while entropy is between $[0, \infty]$.

\subsubsection{The Difference Between our Framework and Calibration and Typical Epistemic Uncertainty Estimations}
Note that our Bayesian Evaluation framework is intended for evaluation of predictors only and requires an accepted label set as per the norm of supervised learning, which is obtained from annotator aggregation methods in subjective annotation tasks.
Measures of calibration and epistemic uncertainty at test time both want to know the certainty of a prediction on unobserved data given the prior data.

There is also a difference in estimation of uncertainty approach that affects the practical results.
Calibration oriented methods are more specific to the input $x$ and attempt to model the uncertainty in the transformation from input space to label space.
Our framework's distribution tends to be simpler to model as it does not rely upon the often complicated input space $X$, which in computer vision is all possible images, and instead only needs to model the possible transformations of the predictions $\hat{y}$ to the accepted label $y$ within the same label space, which avoids modeling the transformation from input to output space and may be done so without knowledge of the internals of the predictor.

Furthermore, our approach does not rely on any assumptions of convenience over the parameters of the predictor's model, which is not a black-box uncertainty estimation, nor any assumption of convenience in the distributions of parameters in our BNN instance as a Bayesian Evaluator.
Most calibration and prior epistemic uncertainty estimation works rely on assumptions of these parameters, such as a Gaussian distribution per parameter, that have no theoretical backing, but do provide more efficient computation times.
Our distribution over the parameters starts with a uniform prior and no further assumptions, which respects the Maximum Entropy Principle of Jaynes et al. 2003~\cite{jaynes_probability_2003}.
Respecting this Maximum Entropy Principle means that as long as convergence occurs in the MCMC sampling method, as it does for all of our experiments, then the resulting distribution is the actual distribution of the parameters in the model (our BNN) given the data.

    \subsection{The Perfect Predictor}
        In the case of a perfect predictor for a task, the situation exists that given the data sampling $(x_i, y_i)$, the predictor $\hat{f}(\cdot)$ perfectly outputs $\hat{y}$, which is the expected task label $y$, thus $\hat{y} = y$.
        The perfect predictor thus has no error and thus has no epistemic uncertainty, because the predictor is not causing any uncertainty whatsoever in its outputs.
        The perfect predictor's possible mappings of the measurable spaces $E_{Y} \rightarrow E_{\hat{Y}}$ all must then either be the identity function or achieve the same results as the identity function on the data sampling $(x_i, y_i)$.
        With $\hat{Y}$ as defined in Sec.~3 of the main text, in the case of a perfect predictor, $\hat{Y} = Y$ and $P_{\hat{Y}}(\hat{y} | y) = P_Y(y | \hat{y}) = 1$ for all pairs $(y, \hat{y})$ based on the given data.
        The Bayesian Evalutor would then simply be the identity function or any function that performs similarly to the identity function on the given data, when using a prior that satisfies the principle of maximum entropy~\cite{jaynes_probability_2003}.
        Of course, all of this assumes that this is with enough data that is representative of the general task being learned.
        
        If this to were occur in practice, the skeptical data scientist would rerun their experiment on a different seed at least once if not multiple times.
        If all outcomes always resulted in a perfect predictor and the predictor is functioning as expected without cheating or bugs, then the data scientist can be empirically certain that this predictor is perfect given the data sampling $(x_i, y_i)$.
        As is well known in machine learning literature, that task given that data would be considered solved, and that in order to better assess the predictor, more data would be required to ensure that the general task has indeed been learned.
        
        Note, however, that in the terms of Shannon information entropy there could still be an expected uncertainty in the task label $Y$ that is carried through to $\hat{Y}$.
        This is the mix of the aleatoric uncertainty --- noise due to the data --- and inherent uncertainty from the random phenomena as defined by Shannon information entropy~\cite{shannon_mathematical_1948} (\textit{e.g.}, a perfect die still has uncertainty as defined by the probabilities of its outcomes).
        Aleatoric uncertainty tends to be defined as noise in the data due to the observation sampling process, \textit{e.g.}, sensor 
        noise~\cite{abdar_review_2021}.
        Given the setup of supervised learning where $Y$ is the random variable of the task labels that defines the task in conjunction with the input data's random variable $X$, in the traditional supervised learning setup and in any scientific experiment the quality of the model learned or evaluated depends on the data.
        This paper does not focus on aleatoric uncertainty.
        The point being made is that even though there is a perfect predictor, the task labels themselves, and thus the predictors perfectly matching output, may still have uncertainty, but \textit{not} epistemic uncertainty in this predictor given the data $(x_i, y_i)$.
        This is related to the main text's Sec. 2 on Task Relevant Information, SM~Sec.~\ref{sec:sm:info}, and SM~Sec.~\ref{sec:sm:info_conserve}.

\subsection{Instances of the Bayesian Evaluator}
The following is the supplementary material for the two instances of the Bayesian Evaluators examined in the experiments in this paper.
Remember that any Bayesian Evaluator may be used, not just these two specific instances.
However, it is important to use a Bayesian Evaluator that has statistical properties that promise the appropriate sampling of the posterior distribution $P_{\hat{Y}}(\hat{Y}|Y)$, as the BNN does using HMC or any MCMC fitting method.
NDoD relies on MLE estimates of the mean and covariances from data, and does not promise any such properties as promised by MCMC-based methods.

\subsubsection{Transformation of the Label Space in the Implementations of the Proposed Framework}
    \label{sec:transform}
    In this work, both Bayesian Evaluators rely on a transformation of the label space to model the distributions within the $K-1$ probability simplex.
    The transformation removes an unnecessary dimension from the existing $K$ probability vector space via a rotation of the space that zeroes out an arbitrary dimension.
    The $\Delta^{K-1}$ space is always able to be expressed in $K-1$ dimensions.
    For example, in the case of a classifier for three classes $K=3$, the $2$-simplex is a triangle in 3D-space whose vertices are one hot vectors able to be expressed as a $3 \times 3$ identity matrix.
    To depict a point on the surface of this triangle, only two dimensions are needed, so a rotation of the triangle can drop the unnecessary 3rd dimension.
    This transformation of the space and its inverse is achieved in this work by translating the probability simplex such that one of its vertices are moved to the origin of the $K$-dimensional space.
    A rotation matrix $Q$ is found via QR factorization of the following rectangular matrix,
   
    \begin{center}
    $\begin{vmatrix}
        -1 & -1 & \dots & -1 & -1 \\
        1 & 0 & \dots &  & 0 \\
        0 & 1 & & &  \\
        \vdots & & \ddots & & \vdots \\
        & & & 1 & 0  \\
        0 &  & \dots & 0 & 1 \\
    \end{vmatrix}$
    \end{center}
    which is a vector of $K-1$ negative ones on top of the $K-1 \times K-1$ identity matrix.
    
    Using the rotation matrix $Q$, the $K-1$ simplex is rotated about the origin such that the unnecessary dimension of the space is zeroed out.
    The inverse rotation followed by the inverse translation allows the space to be transformed back into its original $K$-dimensional space.
    Moving down into the probability simplex space allows for statistical models to be more easily applied, albeit their sample space still needs to respect the boundaries of the transformed simplex, otherwise they will have invalid samples. 
    In this work, the statistical model defined in Sec.~3 of the main text that models the distribution over the differences between the human frequency and the predictor as a multivariate Normal distribution (Normal Distribution Over the Differences (NDoD)), relies on resampling to preserve valid probability vectors as samples.
    The Bayesian Neural Network uses the softmax activation to ensure that its outputs are always valid probability vectors avoiding costly resampling.

\subsubsection{Bayesian Neural Network Implementation Details}
\label{sec:sm:bnn}
    This section elaborates more on the BNN implementation details discussed in Sec.~3 of the main text ``Bayesian Neural Network''.
    The network architecture of the BNN is shown in SM~Table~\ref{tab:bnn}.
    The hidden layer's number of units is selected to be $K-1$ dimensions, the same as the necessary dimensions to describe the subspace of the $K-1$ probability simplex.
    The transformation before and after the BNN's dense layers is not absolutely necessary for the BNN's use as the softmax function would ensure that the output is always a valid probability vector.
    However, the removal of the unnecessary dimension is intended to decrease the parameter space of the BNN to the minimum without forming a bottleneck.
    By limiting the number of parameters in the BNN, the BNN is less likely to be biased on the data, which is a desirable characteristic and a quality that is assessed in Experiment~1.
    However, it may occur in some situations that wider layers or deeper layers are necessary for the specific problem.
    
    The BNN is designed to create Gaussians around the predictions to account for the variance of the predictor given the human frequencies.
    When only one prediction per task sample is available, the hyperparameter will need to be tuned or set to a value that matches the lower bound of expected variance per prediction.
    If resources allow, this decision can be further informed by training multiple predictors on the same training set with different initialization and using their outputs as a set of predictions per task sample.
    
    \begin{table*}[h]
    \begin{center}
    \scalebox{.85}{
    \begin{tabular}{|c|}
        \hline
        input layer: $y$ \\
        \hline
        Probability Simplex Rotation: $-1$ Dim \\
        \hline
        Dense layer: $|y|-1$ units\\
        \hline
        Sigmoid \\
        \hline
        Dense layer: $|y|-1$ units\\
        \hline
        Probability Simplex Rotation: $+1$ Dim \\
        \hline
        Softmax \\
        \hline
        \hline
        Target Likelihood: 
        $\Like(\psi|y, \hat{y}, \sigma) = \N(\text{BNN}(y, \psi) - \hat{y}, ~ \sigma^{2}\text{I})$
        \\
        \hline
    \end{tabular}
    }
    \end{center}
    \caption{
        The proposed Bayesian Neural Network's architecture.
        Similar to Bayesian Regression, the target likelihood function is a multivariate normal whose mean is the difference between the BNN output and the predictor's output $\hat{y}$.
        The normal distribution's covariance matrix $\sigma^{2}I$ is a hyperparameter of the model that sets the lower bound of the variance of the BNN.
        See SM~Sec.\ref{sec:sm:bnn}.
    }
    \label{tab:bnn}
    \end{table*}
    
    To obtain the posterior distribution using the BNN, HMC~\cite{duane_hybrid_1987} is used in this work to sample over the weights and biases of the network.
    The use of softmax in the BNN is intended to respect the probability simplex bounds, and the BNN weights provided by a converged HMC will ensure they are from $P_{\hat{Y}}(\hat{Y} | Y)$.
    The parameters $\psi$ sampled by HMC use the same HMC chain, meaning they are all given the same step size.
    Each sample of the HMC results in a complete weight set of the neural network, which is used to feed the target labels $y$ forward through the network to obtain one sample of $P_{\hat{Y}}(\hat{y}|y)$.
    After reaching convergence and selecting the appropriate lag using the autocorrelation function to overcome the autocorrelation of the HMC chain, multiple draws from the chain are obtained and the input target labels $y$ are fed forward through the BNN with each weight set.
    This creates a tensor with the shape $(N, B, K)$, where $N$ is the number of task samples, $B$ is the number of drawn weight sets from the BNN, and $K$ is the dimensionality of the probability vector. 
    Note that this resulting tensor shape is the same for all Bayesian Evaluators, including NDoD and the others in Experiment 1.
    
    The number of leap frog steps used in HMC was 3 and this remained constant for all sample chains.
    In order to reach convergence, multiple HMC chains were initialized at varying step sizes, all of which were randomly selected at first and updated using a simple gradient based search that increased or decreased the step size as necessary to obtain the desired acceptance rate of 80\%.
    Multiple step sizes were tested at the same time using different chains that were run in parallel in order to expedite the search for the optimal step size given the desired acceptance rate.
    Also, the multiple chains allowed for better detection of convergence when the chains oscillated about the same value for the target likelihood function on the given data.
    HMC chains were considered to have converged once the desired step size was achieved, multiple chains had oscillated about the same value of the log of the target likelihood function, and the multiple individual chains' recorded log likelihood sequences each had a slope less than 1E-7 over the last 1E6 samples drawn.
    After convergence, the HMC chains were sampled from in parallel using the burn in and lag equivalent to the value found from the autocorrelation function to avoid autocorrelation in the samples of the HMC chains.
    
    The convergence process is the longest part of fitting the BNN, followed by sampling the converged chains, but the sampling process is trivially parallelizable.
    Over all the datasets, the approximate time for HMC chain convergence was 3-4 days.
    There were some extreme cases that took 5-6 days, and others that took only a day to converge.
    The runtime of the convergence and sampling process depends highly on the amount of data in the dataset and the predictor's outputs.
    The BNN's hyperparameters also affect the convergence and sampling rates.
    
The BNN's target likelihood function is that of a multivariate normal about the differences between the output of the BNN and the predictor's output $\hat{y}$
\begin{equation}
    \Like(\psi|y, \hat{y}, \sigma) = \N(\text{BNN}(y, \psi) - \hat{y}, ~ \sigma^{2}\text{I})
\end{equation}
where $\sigma$ is a hyperparameter that stands for the standard deviation, I is the $K \times K$ identity matrix, and their product serves as the covariance of the multivariate normal.
The standard deviation hyperparameter in this case determines the lower bound of the variance of the BNN's output per predictor's output.
As the lower bound, the standard deviation can be used to increase or decrease the lower bound of the evaluator's variance.
    Notably, as the value of $\sigma$ decreases, the HMC chains took a longer time to converge.
    In the experiments for this paper, the hyperparameter $\sigma^{2} = 0.1$ is used.
    
    The target likelihood function of the BNN is similar to Bayesian regression.
    The use of this likelihood results in the assumption that the variance of the individual predictor outputs follow a multivariate normal distribution, similar to kernel density estimation with a Gaussian kernel.
    The covariance hyperparameter is similar to the bandwidth of kernel density estimation in how it affects the resulting distribution estimation.
    However, the BNN is able to have varying covariances for its input $y \in Y$, while typical KDE with a Gaussian kernel uses the same bandwidth parameter for all of its Gaussians.
    Using a softmax activation on the output of the BNN avoids costly resampling and respects the bounds of the probability simplex $\Delta^{K-1}$, while using kernel density estimation or NDoD do not respect the bounds of the probability simplex and require resampling of invalid samples.
    
    When there are multiple subsets of data that may have different conditional distributions $P_{\hat{Y}}(\hat{Y} | Y)$, such as the training and testing set of a predictor, fitting a single BNN to all of the data may not easily capture the differences between the subsets.
    In fact, fitting one BNN to all of the data will only allow for modeling the entire data's mixed conditional distributions, as it is designed to do by fitting the BNN in such a way.
    An individual BNN for the predictor's training and testing sets is recommended when $P_{\hat{Y}}(\hat{Y} | Y)$ has the potential to be different between the two sets, \textit{e.g.}, overfit predictors. 
    This results in a mixture of two BNNs that allows for analyzing the conditional distribution of each subset individually, while also allowing the overall conditional distribution to be analyzed by sampling from each BNN using the ratios of the amounts of data within each subset.
    This individual BNN fitting for the training and testing sets is done in Experiment 2 and 3 in the main text.
    
    As mentioned in Secs.~1--2 of the main text, this evaluation framework is intended to assess predictor uncertainty where the predictors are treated as black boxes.
    The BNN used as the Bayesian Evaluator as described in this paper is intended as a more general model, and when used generally has uniform priors over the weights and biases of the neural network.
    This respects the principle of maximum entropy~\cite{ jaynes_probability_2003} by making no assumptions over the possible distributions of the weights and biases in the network.
    This is in contrast with other works that rely on the Gaussian prior distribution on the weights and biases as noted in Sec.~2 of the main text.
    The benefit of this evaluation framework being able to be applied to any predictor comes at the cost of needing to have enough data to be able properly estimate the conditional distribution.
    This is the case in all statistical approximation methods, but it is still worth reminding any potential users of this situation.
    In the typical supervised learning case for training predictors, this assumption of enough data to adequately define the general task is already made, which implies that there is also enough data to approximate the conditional distribution $P_{\hat{Y}}(\hat{Y} | Y)$.
    
    A good Bayesian Evaluator must be able to approximate the possible transformations of $E_{Y} \rightarrow E_{\hat{Y}}$ not only for the data samples provided, but also for data samples not provided.
    Experiment 1 in the main text tests for this by splitting the data in half and checking how the Bayesian Evaluator fits across the halves, when fit on only one.
    NDoD approximates this by assuming that all inputs have the same Normal distributed error, which holds when the predictor is performing well enough to have its estimates around the expected label.
    Thus, NDoD applies this error to any given input and accomplishes its approximation of possible transformations given the data and its model this way.
    The BNN is more nuanced where it models both the bias and covariance of the Normal distribution to define the possible transformations for \textit{each} target label.
    In the sample space of probability vectors, this means that the BNN approximates the potential mappings of a given task label to the possible values of the predictor over the entire probability simplex, given the data and the BNN's own architecture.
    Thus the BNN is a fairly capable model of $P_{\hat{Y}}(\hat{Y} | Y)$, given enough data and MCMC convergence.

\section{Supplemental Material for the Experiments}
\label{sec:sm_exps}
The supplemental material below provides the details necessary to reproduce the experiments described in Sec.~4 of the main text, as well as provide supplemental experiments to further evaluate the fit of the Bayesian Evaluators and depict the use of the BNN in practice.

\begin{table*}
\centering
\scalebox{0.82}{
{\setlength{\tabcolsep}{0.41em}
    \begin{tabular}{
        llll
        l@{\hspace{0.4em}}l@{\hspace{0.4em}}l@{\hspace{0.4em}}l@{\hspace{0.4em}}l
        l@{\hspace{0.4em}}l@{\hspace{0.4em}}l@{\hspace{0.4em}}l@{\hspace{0.4em}}l
    }
        Dataset & Class & Samples & Total & \multicolumn{5}{c}{Humans / Sample} & \multicolumn{5}{c}{Labels / Human} \\
        & Bins &  & Humans
            & m & Q1 & Q2 & Q3 & M
            & m & Q1 & Q2 & Q3 & M \\
        \hline
        LabelMe\cite{rodrigues_learning_2017}
            & 8 & 1000 & 59
            & 9 & 10 & 11 & 11 & 11
            & 3 & 9 & 27 & 66 & 182 \\
        SCUT-FBP5500\cite{liang_scut-fbp5500:_2018}
            & 5 & 5500 & 60
            & 60 & 60 & 60 & 60 & 60
            & \multicolumn{5}{c}{5500} \\
        APPA-REAL\cite{agustsson2017appareal}
            & 20 & 7591 & N/A
            & 10 & 14 & 37 & 39 & 609
            & \multicolumn{5}{c}{N/A} \\
        TestMyBrain: Trust 
            & 7 & 6898 & 5327 
            & 12 & 26 & 31 & 52 & 94
            & 46 & 49 & 50 & 50 & 50 \\
        \cite{mccurrie2017predicting, germine2012web}
    \end{tabular}
}}
\caption{
    The crowd annotated computer vision datasets used in the experiments.
    Class bins is the total classes of the dataset.
    APPA-REAL's dataset was discretized into bins of 5 years of age.
    The last two columns are the minimum, quartiles, and maximum of the unique human annotators per sample and the annotations per human annotator.
    This depicts the annotation coverage of the task samples and indicates the overlap of the humans' annotations. 
    LabelMe's total number of humans is after the removal of annotators who did not provide any annotations, possibly artifacts due to not completing the survey.
    APPA-REAL did not track the unique annotators' responses which resulted in not having a total number of unique human annotators nor knowing the number of labels completed by each human. 
    SCUT-FBP5500 had every annotator annotate every sample, this is why ``Labels / Human'' column is 5500.
}
\label{tab:data}
\end{table*}

\subsection{Dataset Details and Use}
    Details for each dataset are provided in SM~Table~\ref{tab:data}.
    The datasets were mostly used as intended, with the exception of LabelMe, whose only difference in use was in the removal of annotators whose responses were invalid values, indicating they did not provide any annotations.
    This could have occurred for multiple reasons, such as the annotators not finishing the survey or somehow providing invalid annotation labels.
    This change is notable because with their removal the dataset statistics change slightly from the original work's published statistics.
    
    The dataset statistics in SM~Table~\ref{tab:data} indicate that the four examined datasets are of varying quality with LabelMe having the least coverage of different humans per samples, which results in the human frequencies having coarse granularity.
    The ability to express subtle variations in the probability vector labels is what is meant by granularity for human frequency.
    Fewer human annotators per sample results in less smoothness in the probability vectors values.
    This results in clustering of the resulting human frequencies when subtle nuances in the probability vectors for the samples may better represent of human behavior.
    This lower amount of human annotators per samples may be the cause of the multimodal measure distributions over normalized Euclidean distances and KL Divergences in this paper's experiments.
    APPA-REAL has the largest maximum annotators per sample and also the second lowest minimum, which indicates that the resulting human frequencies vary greatly in the amount of granularity.
    SCUT-FBP5500 is the best covered dataset with all of its annotators having annotated every sample.
    
    The implementation of the predictors followed their respective papers' recommendations and the same for the datasets' samples.
    The only data preprocessing required for the datasets was a resizing operation to fit their respective predictor inputs and color normalization.

\subsection{Predictor Training Details}

    For LabelMe, a VGG16 network was used whose weights were frozen on ImageNet and whose final layer was replaced with a dense layer of 128 hidden units with a ReLU activation followed by a drop out of 50\% of the units.
    The output layer of this network was a dense layer of the class size followed by the softmax function.
    For SCUT-FBP5500, a ResNeXt50 
    network was initialized with ImageNet weights, but not frozen. 
    The LabelMe and SCUT-FBP5500 classifiers were trained with early stopping informed by the testing set with a patience of 5 epochs and taking the best weights.
    For both ``TestMyBrain: Trustworthiness'' and APPA-REAL, a ResNet-18 pretrained on Imagenet was trained for about 100 epochs, similarly replacing the final softmax layer with a single fully connected layer. 
    
    \subsubsection{Computational Resources}
    The training of the predictors was the only case where GPUs were used in this work.     Only one GPU was used to train a predictor at a time.
    The rest of the calculations, specifically the HMC chains' convergence and sampling, were all accomplished on CPUs using the Tensorflow Probability API.
    A shared cluster was used for this.
    Each chain during convergence took only 1 CPU, but given the way the compute cluster was configured, the amount of RAM was associated with the number of CPUs assigned to the job, and typically 4 CPUs were assigned for an individual HMC chain, giving it 40 GB of RAM to allow for loading the task data and predictor's outputs, but leaving space for sampling the chain. 
    The Bayesian Evaluators were sampled approximately 14,000 times, resulting in 14,000 sets of predictions.
    In the case of the BNN, this was 14,000 sets of weights and biases used to obtain the corresponding predictions.
    The mean number of CPUs used for the sampling of the HMC chains was 4 CPUS per 5 job submissions for further parallelization, resulting in 20 CPUs used on average for HMC chain sampling.
    The large number of machines and CPUs available expedited the running of all experiments by allowing them to be parallelized.
    
    Given these resources, the individual predictors trained fairly fast, lasting only a day at most.
    The varying use of the compute cluster by others resulted in variance in the wall time of the different computations, but the HMC chains tended to complete converging on average in 3 days.
    Sampling at longest took a day, but tended to only take 4 hours on average per dataset.
    The datasets' sample sizes did result in different runtimes.
    The larger datasets took longer for the HMC chains to run due to having to calculate the sum of the log of the target liklihood function for the BNN on all of the data for every sampling of the BNN's weights and biases.
    The visualizations took a similar amount of time to complete compared to the sampling of the BNN.
    The NDoD fitting and sampling were done sequentially and were either very fast relative to the rest of the computations or very slow.
    The fast instances were for the SCUT-FBP5500 and ``TestMyBrain: Trustworthiness'' datasets, while the slower instances were on LabelMe and APPA-REAL.
    As noted in the main text, APPA-REAL's NDoD models never finished resampling.

\subsection{Measure Details: Normalized Euclidean Distance, Kullback-Leibler Divergence, and Area Under the Curve}    
\label{sec:euclid}
Given that the probability simplex is a regular simplex with equal length sides and the vertices expressed as one hot vectors, the maximum Euclidean distance is then the magnitude of each simplex edge $\sqrt{2}$, which may serve as a normalizing constant of the Euclidean distance.
This normalizing constant simply rescales the Euclidean distance so the range of possible values is $[0,1]$ instead of $[0, \sqrt{2}]$ such that the interpretation is more similar to other bounded measures of similarity and is more user friendly in analysis.
\begin{equation}
    L_2(a,b) = \frac{1}{\sqrt{2}}||a - b||^{2}_{2}
\end{equation}
When Euclidean distance is applied to a reference point (human frequency $y$) and a point in question (prediction $\hat{y}$), it captures how much the point in question differs from the reference point.
When applied to all pairs of points, the distribution of Euclidean distances expresses how much the distribution in question differs from the reference distribution.
This makes the distribution of Euclidean distances a sufficient statistic for assessing the fit of the predictors to their target labels and of Bayesian Evaluators of $P_{\hat{Y}}(\hat{Y}|Y)$ to the actual predictor's outputs.
    
As the proposed framework is agnostic of the measure used, the measures examined other than the normalized Euclidean distance included the Kullback-Leibler (KL) divergence applied to each sample target label to prediction, which are both probability vectors:
\begin{equation}
    D_{KL}(P, Q) = \sum_{v \in V}P(v)\log_2\left(\frac{Q(v)}{P(v)}\right)
\end{equation}

The area under the curve of the receiver operating characteristic (AUC) is calculated by applying the argmax function to the classifier's output and the human frequencies.
The KL divergence and AUC results are included below.

\subsection{Experiment 1: Supplemental Experiments}
The following are the supplemental details and experiments to further assess the fit of the Bayesian Evaluators in approximating $P_{\hat{Y}}(\hat{Y} | Y)$.
    To assess the Bayesian Evaluators' fit and generalization, each predictor's training and testing sets were combined, shuffled, and then split in half to create the training and testing sets for the Bayesian Evaluators of $P_{\hat{Y}}(\hat{Y} | Y)$ for all of Experiment 1.
    The reason the predictors' original training and testing sets were not used is because of the possibility of the predictor behaving differently on the two sets due to under or over fitting.
    If the predictor behaves differently on the two training and testing sets, then that means there are two different distributions that need to be learned.
    The combination and shuffle of the sets avoids this issue and simplifies evaluation of the Bayesian Evaluators.
    
    This experiment assesses only the Bayesian Evalutors, where different models are compared to one another by how well they represent the predictor's outputs.
    By combining the predictor's training and testing sets, the experiment's variables are better controlled, the bias and variance of the Bayesian Evaluators can be compared, and excess bias or variance in the individual models can be detected.
    Note that because the Bayesian Evaluators are only operating in an evaluation context, it is fair to combine these sets.
    
    Regarding the second baseline Bayesian Evaluator used in Experiment 1 that assumes independence between $Y$ and $\hat{Y}$, the Dirichlet distribution with alternate parameterization~\cite{minka2000estimating} had its parameters selected in a specific way.
    This alternate parameterization specifies a Dirichlet distirbution by the mean and the precision, where the mean is simply the means along each element in the probability vector and the precision determines how concentrated the density is about that mean within the Dirichlet's label space.
    Another way to think of this parameterization is that the mean determines the frequency or ratio and the precision parameter determines the intensity of the mean for the Dirichlet distribution.
    This Dirichlet distribution with alternate parameterization used the uniformly minimum-variance unbiased estimator of the mean of the data as a constant parameter.
    Gradient descent via the ADAM optimizer was used with the precision parameter as the lone free variable changed to maximize the log likelihood of the data being drawn from the Dirichlet distribution, finding the maximum likely parameterization.
    
    Notably, the NDoD and Zero Mean NDoD both were unable to finish sampling for the APPA-REAL dataset within the timeframe of this work (unable to sample even once within 30 days).
    This is to be expected, and was  noted in Sec.~3 of the main text when describing NDoD.
    NDoD was intended to serve as a simple Bayesian model whose design makes sense for predictors who are correct most of the time with some degree of error.
    In other words, NDoD is for predictors whose output tends to center around the task label.
    However, designing statistical models of conditional distributions that are bounded to a simplex is a non-trivial task, and is not as simple as using a truncated distribution or Dirichlet distribution.
    The simplicity of NDoD is desirable for quick implementation as long as the design assumptions hold, but this simplicity comes at the cost of a large increase in resampling rates as the number of dimensions increases.

    \subsubsection{Simulation Experiments}
    \label{sec:sim}
\begin{figure*}[h]
            \centering
            \includegraphics[width=0.8\textwidth]{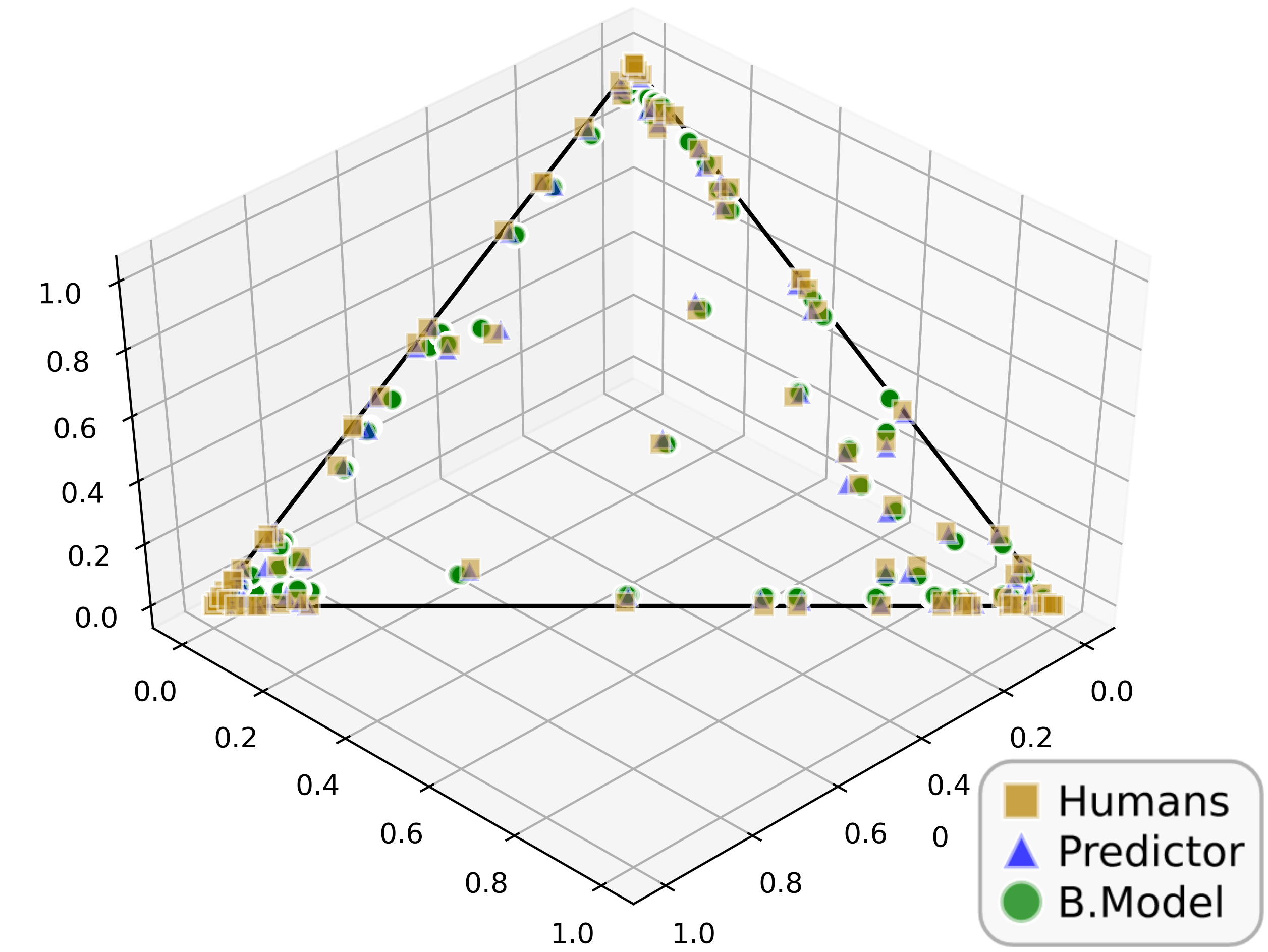}
            \caption{
                An example visualization of the simulation of humans that are certain on the labels of some task, as indicated by higher densities towards the vertices of the probability simplex, and of a simulated predictor that is the identity function with a small amount of Gaussian noise.
                This figure also serves as a useful visual of the probability simplex in 3 dimensional space where the simplex is a triangle.
                Labels that occur towards the center of the probability simplex indicate a more equal probability of the 3 classes indicating no significant favoring of any single class.
                The transformation in SM~Sec.~\ref{sec:transform} in the 3 dimensional case would be the rotation of one of the vertices such that the triangle is parallel to one of the hyperplanes formed by the remaining two dimensions.
                The coordinates for the valid points within the probability simplex would then have zeros in the zeroed out dimension.
            }
            \label{fig:certainSim}
        \end{figure*}

\begin{figure*}
            \centering
            \includegraphics[width=\textwidth]{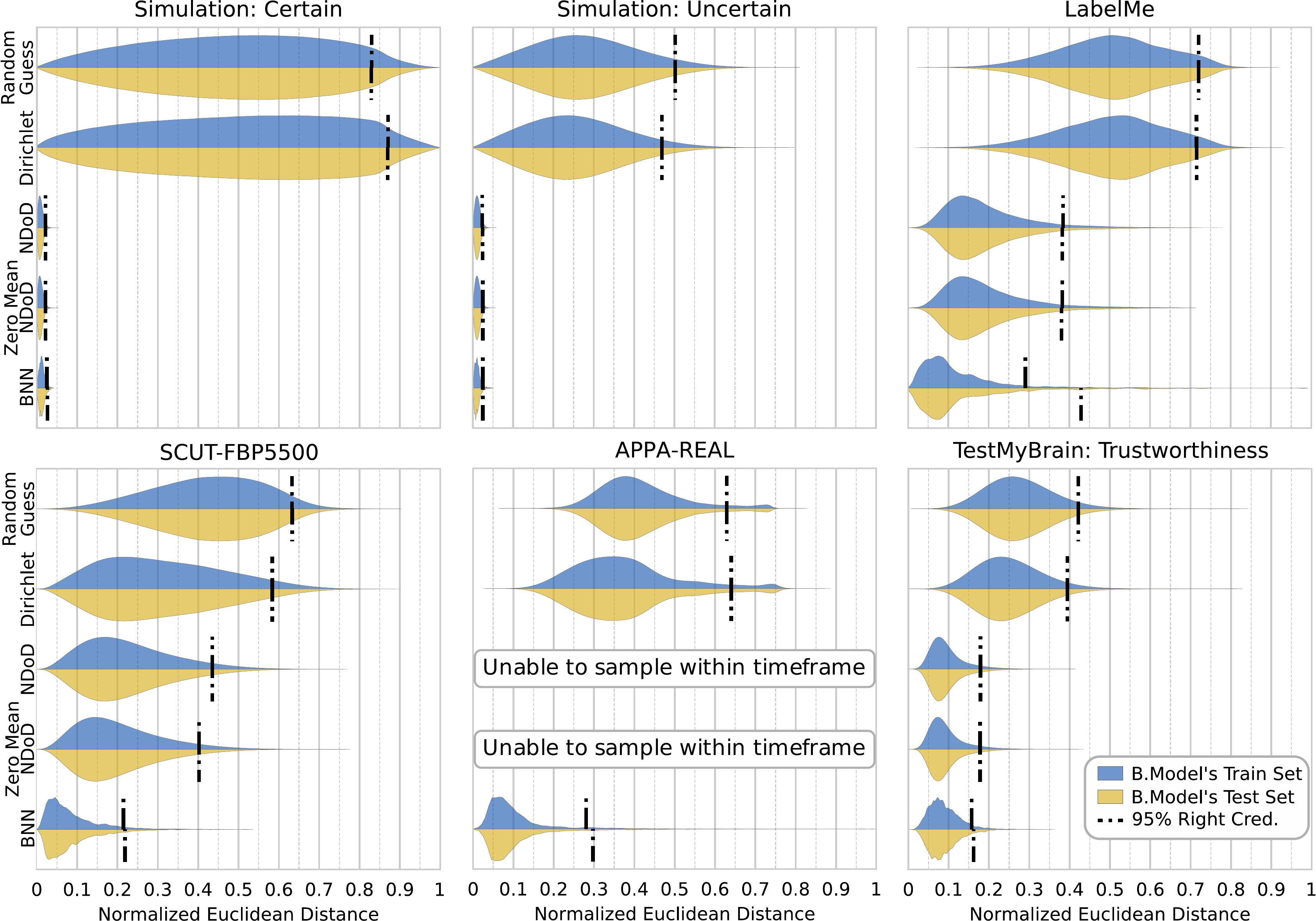}
            \caption{
                Experiment 1's results for how well each Bayesian Evaluator fits the conditional distribution $P_{\hat{Y}}(\hat{Y} | Y)$ of the predictor output given the human frequency by measuring the normalized Euclidean distances between the models' samples to the predictor's output.
                This figure is the same as Fig.~3 in the main text, except it also includes the results of the simulations from SM~Sec.~\ref{sec:sim}.
                The simulations are of a uniform Dirichlet with 0.2 as all concentration parameters to model a situation where the humans are certain on the task, and of a uniform Dirichlet with 10 as all concentration parameters to represent when the humans are uncertain.
            }
            \label{fig:exp1_sim}
        \end{figure*}

        In addition to the four datasets, two simulated datasets were used in the experiment in order to assess how well NDoD and the BNN serve at approximating the conditional distribution $p(\hat{y} | y)$ in two different scenarios of a well-trained predictor.
        The two simulated scenarios include when the human frequencies are overall uncertain and certain of the task, but the predictor matches the human frequencies well with a small amount of Gaussian noise.
        The simulations were built to model the 3-dimensional probability vector label space.
        Both of these simulations modeled the predictor as an identity function with a small amount of Gaussian noise, $\N (0, (1e-4)I)$, where $I$ is the identity function.
        These simulated situations were to provide a trivial case for the NDoD and the BNN, as the simulation matches how the NDoD models the problem, as seen in Eq.~7 of the main text.
        Using the same simulation of the predictor, the two simulations modeled the target labels as Dirichlet distributions with concentration parameters equal to ${(10, 10, 10)}$ and ${(0.2, 0.2, 0.2)}$.
        These Dirichlet distributions simulated the situation where the majority of humans were uncertain and certain of all of the classes, respectively.
        The simulation training and testing sets for this experiment were both 1000 samples.
        SM~Fig.~\ref{fig:certainSim} is a visualization of the simulation of certain humans. SM~Fig.~\ref{fig:exp1_sim} shows the complete results for the simulated scenarios, along with the Experiment~1 results from the main text for comparison.
        The NDoD models and the BNN fit both simulations nearly perfectly, indicating that they are working as intended and that the BNN is capable of fitting a slightly noisy identity function equivalently to the NDoD models, which is the intended scenario for NDoD.
        The BNN model for ``Simulation: Certain'' deviates from SM~Table~\ref{tab:bnn} by using 5 hidden units in its one hidden layer instead of 2 units, and both BNNs for the simulations use $\sigma^2 = 1.0$E-3 to allow for learning the lower variance of the predictor.
        
    \subsubsection{Data Sensitivity Assessment via 3 Fold Cross Validation}
\begin{figure*}[h]
        \centering
        \includegraphics[width=\textwidth]{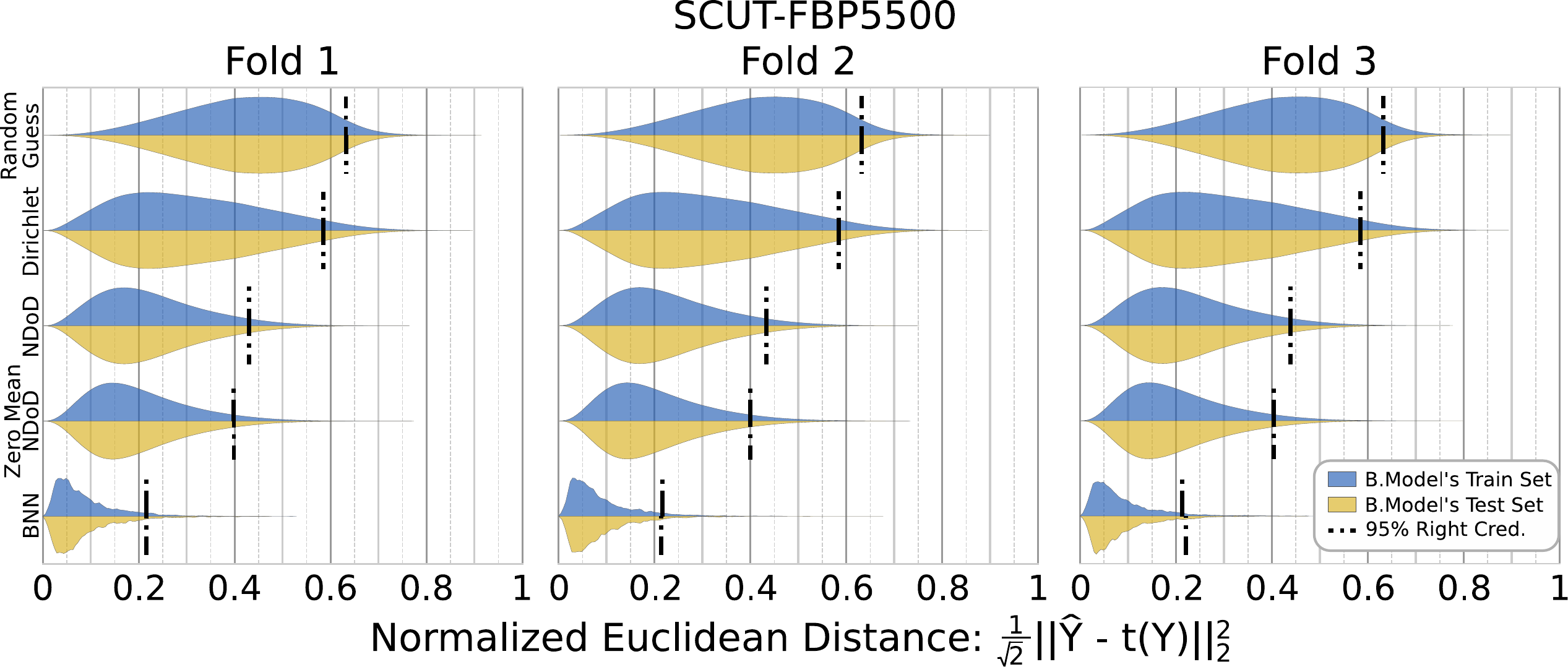}
        \caption{
            The resulting normalized Euclidean distance distributions of the predictor's output to the different Bayesian models using 3 Fold Cross Validation on the  predictor's training and testing sets.
            Given that this is 3 fold cross validation, each Bayesian model is fit on $2/3$ of the data and tested on $1/3$.
            This figure's format is exactly the same as that of the main text's Fig.~3 and SM~Fig.~\ref{fig:exp1_sim}.
            A difference between those two figures is that those Bayesian models are fit on one half of the data.
            This supplemental experiment is to assess the sensitivity of the different Bayesian models to different partitions of the same data and to examine how it affects their generalization.
            All Bayesian models perform almost identically across each fold, where the most difference is in the BNN, but this variance appears to be functionally equivalent to minimal noise, as the resulting distributions and their 95\% RTCI are all nearly identical.
        }
        \label{fig:exp1_3fold}
    \end{figure*}

    Experiment 1 fits the Bayesian models on half of the available data and compares the training set performance to the testing set performance to compare how the Bayesian Models generalize on unseen data.
    To further examine the generalization of the Bayesian models and their sensitivity to the data, 3 Fold Cross Validation is used on the SCUT-FBP5500 dataset with its respective predictor trained to its termination condition.
    This experiment has all of the predictor's training and testing sets combined in order to represent the entire conditional distribution of the predictor given the target.
    The Bayesian models are fit to each of their training sets for the 3 folds individually and compared to their respective testing sets. The results are shown here in SM Fig.~\ref{fig:exp1_3fold}.
    
\subsection{Experiment 2: Supplemental Experiments} 
This section provides supplemental experiments for Experiment 2 in the main text to better depict the use of the BNN implementation of the Bayesian evaluation framework in practice.

\subsubsection{Additional Measures: Kullback-Leibler Divergence and Area Under the Curve}
\label{sec:sm_exp2}
\begin{figure*}[ht]
        \centering
        \includegraphics[width=\textwidth]{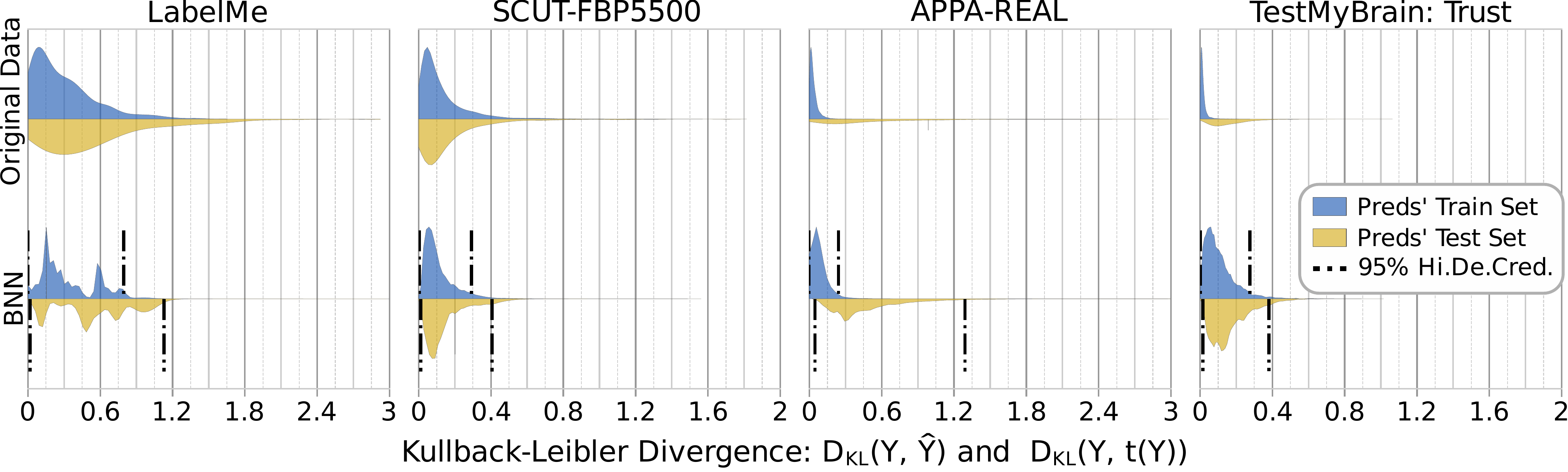}
        \caption{
            The distribution of Kullback-Leibler Divergence of the original measure per sample compared to the BNN pair's flattened samplings of the conditional distribution.
            A BNN per training and testing set was used for this experiment to capture the differences between the two dataset partitions.
            The results are similar to that of the Normalized Euclidean Distances in Fig. 4 of the main text, except that there is no upper bound on the KL Divergence (a shortcoming of this measure for analysis).
            Due to no upperbound on the measure, there were some values above the depicted upper limit of the x-axis, as seen by the long tails of LabelMe and APPA-REAL.
            The chosen KL Divergence upper bounds for the x-axis were chosen to contain as much of the measure distribution as possible, while still allowing the higher densities at the lower values to be visible.
            Similar to the normalized Euclidean distance, the conditional distribution here matches the original measure distribution with the differences seen representing the uncertainty in the predictor given the human frequency. 
            The SCUT-FBP5500 predictor has the most consistent certainty due to the near symmetry across the predictor's training and testing sets and as evident in the 95\% highest density credible interval.
            The second most consistent in certainty is the predictor for ``TestMyBrain: Trustworthiness''.
            Same as with the normalized Euclidean distance distribution, LabelMe's multimodality for KL Divergence indicates the uncertainty of the predictor, possibly due to task difficulty and a lack of data.
        }
        \label{fig:exp2_kld}
    \end{figure*}

The Kullback-Leibler Divergence was examined in the same way as Experiment 2, and the results shown in SM~Fig.~\ref{fig:exp2_kld} are very similar to those of the normalized Euclidean distance in Fig.~4 of the main text.
Given the similarity between the normalized Euclidean distances and the KL Divergences, the Euclidean distance is a more desirable measure as it captures the difference between the human frequencies and the predictor's output (or the BNN's draws from $P_{\hat{Y}}(\hat{Y}|Y)$ using the framework) in an intuitive way.
Also, the normalized Euclidean distance is bounded from zero to one, which makes its values easier to understand as described in SM~Sec.~\ref{sec:euclid}.
The bounds also avoid the possibility of infinite values, which if occur may interfere with the interpretation of the credible interval, as is the case with KL Divergence for APPA-REAL's predictor trained for 100 epochs. These results are shown in SM~Fig.~\ref{fig:exp2_kld}.

    The AUC was also examined, however given that the original measure without using the framework is a single value, the third row of the AUC distribution in SM~Fig.~\ref{fig:exp3_auc} obtained from the BNN corresponds to what would be shown for Experiment 2.
    The original AUC for each predictor across epochs is overlayed on the BNN's approximated AUC distribution.

\subsection{Experiment 3: Supplemental Experiments}
    This section provides further experiments that depict the use of the Bayesian framework with the BNN in practice for aiding in model selection.
    Based on the prior mentioned desired traits for a predictor, the best predictor may be selected for each dataset.
    The best predictor is the one with the best performance as indicated by its measure(s) and with the least uncertainty.
    The balance between the two is determined by the operator for their specific application.
        
    \subsubsection{Additional Measures: Kullback-Leibler Divergence and Area Under the Curve}
        \label{sec:sm_exp3}
\begin{figure*}[h]
        \centering
        \includegraphics[width=\textwidth]{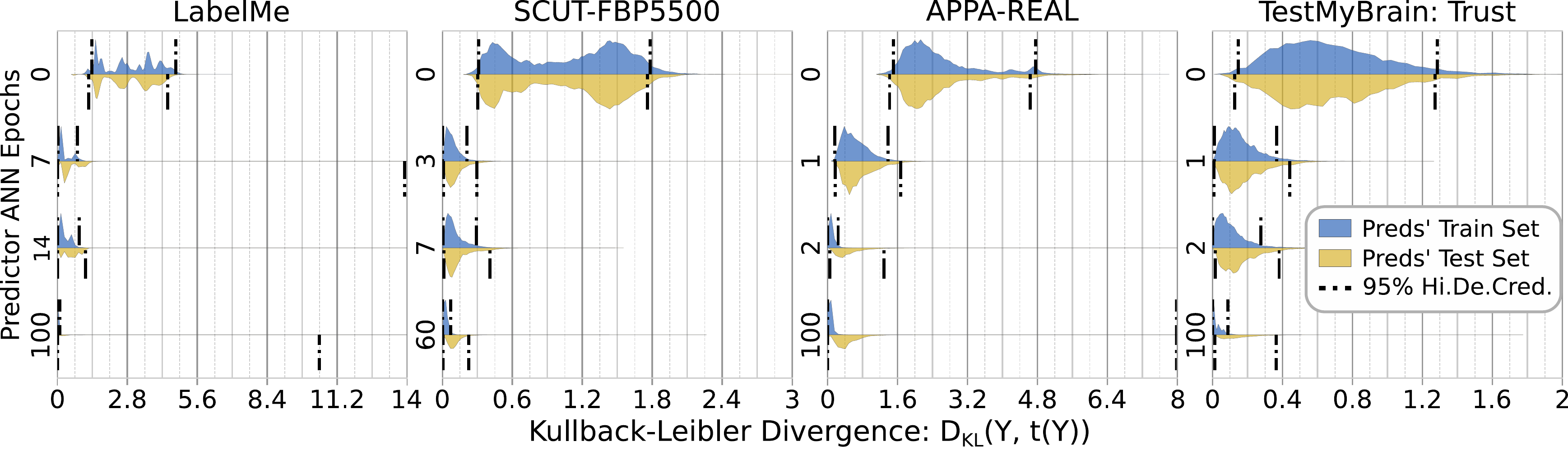}
        \caption{
            The distribution of Kullback-Leibler Divergence approximated by a BNN per training and testing set across the training of the predictors.
            The results are similar to those of the Normalized Euclidean Distances in Fig.~5 of the main text, except that there is no upper bound on the KL Divergence.
            Due to no upper bound, infinitely large values can occur, which is what happened with APPA-REAL's 100 epoch predictor in both its training and testing sets, and results in long tail distributions.
            Given that is the 95\% highest density credible interval, that means that more than 5\% of the measures resulted in infinity large values on both the training and testing sets.
            This measure exhibits the same tendencies and nuances of a predictor as it trains, as seen in the normalized Euclidean distance distributions in Experiment 3 in Fig.~5 of the main text.
            Predictors are the most uncertain when untrained on the task, and improve in performance and certainty as they train over time.
            This trend continues for the training set, but once overfitting occurs, a  predictor's performance worsens and becomes more uncertain.
        }
        \label{fig:exp3_kld}
    \end{figure*}

\begin{figure*}[h]
        \centering
        \includegraphics[width=\textwidth]{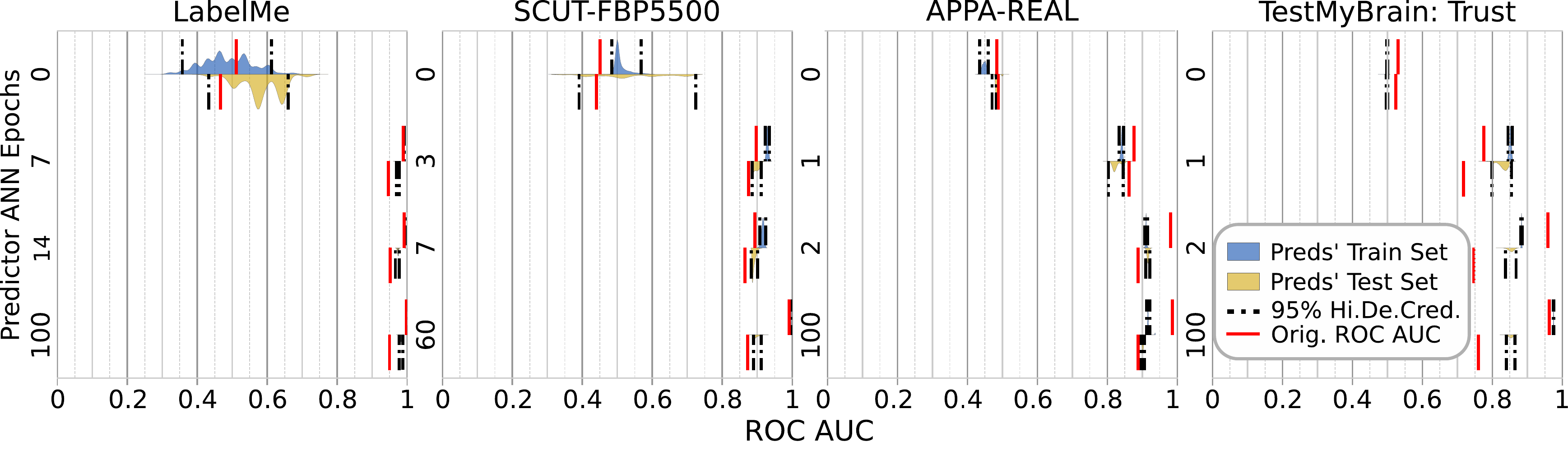}
        \caption{
            The distribution of AUC over training epochs.
            The format here is the same as Fig.~5 of the main text.
            The pair of BNNs, one for both training and testing sets, approximated the AUC distribution.
            Notably, the $P_M(M | Y, \hat{Y})$ tends to be much tighter than the normalized Euclidean distance or KL Divergence distributions.
            This in part due to the AUC being a single point estimate for the entire original dataset, while the other two were a measure per task sample.
            Due to this, the BNNs tended to provide an extremely low variance when the point estimate was more certain.
            The cases where the ROC is most uncertain is when the predictor is untrained, which matches expectations and indicates the uncertainty of $P_M(M | Y, \hat{Y})$ due to the predictor with respect to the human frequencies.
        }
        \label{fig:exp3_auc}
    \end{figure*}

        Experiment 3 from the main paper focuses on the normalized Euclidean distance.
        Here, the same experiment is performed using the BNN per training and testing sets and two different measures: KL Divergence and AUC.
        The KL Divergence results are depicted in SM~Fig.~\ref{fig:exp3_kld} and the AUC results are in SM~Fig.~\ref{fig:exp3_auc}.
        Notably, KL Divergence continues the trend observed in Experiment 2's supplemental experiments, where its characteristics are similar to that of the normalized Euclidean distances.
        
        The AUC on the predictor trained over epochs is in  SM~Fig.~\ref{fig:exp3_auc}, and its third row is the same as the expected experiment result for SM~Sec.~\ref{sec:sm_exp2}, Experiment 2.
        For all experiments where the $P_{\hat{Y}}(\hat{Y}| Y)$ is approximated over the predictor's epochs, the third row is always the same as the BNN's results for the corresponding measure's Experiment 2.
        In this case for AUC they were exactly the same because the original measure was simply added as a red line on the violin plot. 
        
        The original AUC often is outside the BNN's approximated $P_M(M | Y, \hat{Y})$, which potentially indicates two things:
            \textbf{a)} the chosen Bayesian model using the BNN does not represent $P_Y(\hat{Y} | Y)$ well and a new Bayesian model is needed, or
            \textbf{b)} if the BNNs are accurately modeling the $P_Y(\hat{Y} | Y)$, then when the AUC measure is not within the 95\% highest density credible interval, the original AUC is considered significantly uncertain and other measures should be used.
        Due to the possibility of \textbf{a}, it is important to understand that this framework uses approximations of $P_M(M | Y, \hat{Y})$ and that the models that are doing the approximation must be tested against other potential Bayesian evaluators and have their hyperparameters tuned.
        This framework is meant to supplement the original measures, and so the potential for an ill-fitting Bayesian model of $P_M(M | Y, \hat{Y})$ is a possibility that must be handled with care along with the choice of measurements.
        
\end{document}